\pdfoutput=1

\documentclass[11pt]{article}

\usepackage[]{EMNLP2023}

\usepackage{times}
\usepackage{latexsym}
\usepackage{graphicx}

\usepackage[T1]{fontenc}

\usepackage[utf8]{inputenc}

\usepackage{microtype}
\usepackage{dirtytalk}

\usepackage{inconsolata}

\usepackage{lipsum}

\newcommand\blfootnote[1]{%
  \begingroup
  \renewcommand\thefootnote{}\footnote{#1}%
  \addtocounter{footnote}{-1}%
  \endgroup
}

\title{The Sentiment Problem: A Critical Survey towards Deconstructing Sentiment Analysis}

\author{%
Pranav Narayanan Venkit\textsuperscript{$1$}\textsuperscript{*} \enspace Mukund Srinath\textsuperscript{$1$}\textsuperscript{*} \enspace Sanjana Gautam\textsuperscript{$1$} \\ \textbf{\enspace Saranya Venkatraman\textsuperscript{$1$} \enspace Vipul Gupta\textsuperscript{$2$} \enspace Rebecca J. Passonneau\textsuperscript{$2$} \enspace Shomir Wilson\textsuperscript{$1$}} \\
{\textsuperscript{$1$} College of Information Sciences and Technology}  \quad\\
{\textsuperscript{$2$} Department of Computer Science \& Engineering, College of Engineering} \quad  \\
Pennsylvania State University \\
\normalsize{\tt \{pranav.venkit, mukund, sanjana.gautam, saranyav, vkg5164, rjp49, shomir\}@psu.edu} 
}

\begin{document}
\maketitle
\begin{abstract}

We conduct an inquiry into the sociotechnical aspects of sentiment analysis (SA) by critically examining 189 peer-reviewed papers on their applications, models, and datasets. Our investigation stems from the recognition that SA has become an integral component of diverse sociotechnical systems, exerting influence on both social and technical users. By delving into sociological and technological literature on sentiment, we unveil distinct conceptualizations of this term in domains such as finance, government, and medicine. Our study exposes a lack of explicit definitions and frameworks for characterizing sentiment, resulting in potential challenges and biases. To tackle this issue, we propose an ethics sheet encompassing critical inquiries to guide practitioners in ensuring equitable utilization of SA. Our findings underscore the significance of adopting an interdisciplinary approach to defining sentiment in SA and offer a pragmatic solution for its implementation.


\end{abstract}

\section{Introduction}
\blfootnote{* Authors Contributed Equally}

\begin{figure}[h]
  \centering
  \includegraphics[scale = 0.215]{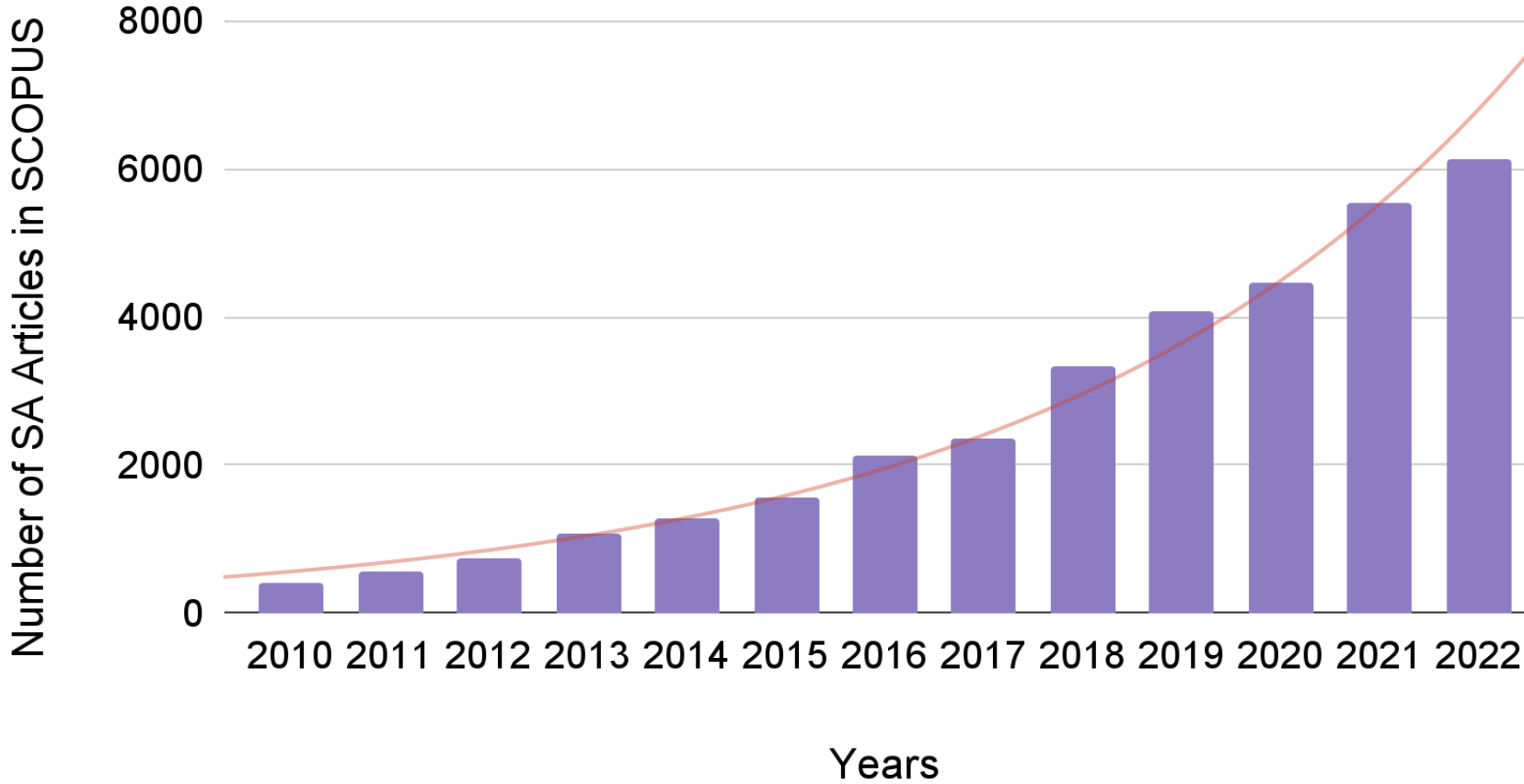}
  \caption{Number of articles published each year (from 2010 to 2022) in SCOPUS that contain the term `sentiment analysis' in the title, abstract, or keywords.}
  \label{fig:scopus}
\end{figure}

Sentiment Analysis (SA) has emerged as a significant research focus in Natural Language Processing (NLP) over the last decade. It has now become an indispensable tool 
in discerning opinions and emotions in written text \cite{medhat2014sentiment}, evaluating social entities' reputation \cite{yuliyanti2017sentiment}, analyzing and predicting financial needs \cite{wang2013financial}, and aiding in effective political decision-making \cite{cardie2006using}. This is illustrated in Figure \ref{fig:scopus} which shows the rising numbers of peer-reviewed articles on sentiment analysis published in SCOPUS every year.

Existing research reveals a notable absence of interdisciplinary endeavors to comprehend the social dimensions of SA, encompassing aspects like emotion and fairness \cite{mohammad2022ethics, blodgett2020language}. This lack of collaborative thinking has resulted in flawed analyses and biased outcomes. Given the extensive range of applications of SA spanning diverse domains such as healthcare, finance, and policymaking, it is crucial to avoid replicating such tendencies. Furthermore, SA, despite addressing social constructs like emotion, subjectivity, and opinion, has been limited in its incorporation of psychological and sociological definitions of sentiment \cite{stark2021ethics}. While numerous studies have examined the utilization of SA, encompassing its inherent challenges and future directions \cite{cardie2006using, zhang2022survey}, the interdisciplinary and sociotechnical dimensions of SA have received limited exploration.


To this end, we explore this gap in the literature by examining sentiment through a technical perspective concentrating on the evolution of SA into a social system. We then evaluate sentiment, examining the various definitions of sentiment through a sociotechnical lens. We also investigate the application of SA, presenting insights into its utilization. These investigations will shed light on the interdisciplinary divide of the term sentiment. Next, we evaluate the motivation behind establishing necessary frameworks for measuring sentiment by examining various different SA models and datasets. 

Through our critical survey of 189 unique works in SA (as shown in Table \ref{table:numbers}, we reveal that very few works (<5\%) in SA try to explicitly define sentiment and sentiment analysis. Our results highlight a lack of effort within the field of NLP to understand the interdisciplinary aspect of sentiment. We also show an absence of synchronization in the field, leading to multiple variations of the term sentiment. Our analysis illustrates how such systems can cause sociodemographic biases due to the lack of nuance in measuring sentiment. To mitigate this issue of an interdisciplinary gap, we propose an ethics sheet \cite{mohammad2022ethics} consisting of ten critical questions to be used as a metaphorical `nutrition label' to understand the issues of SA models by both the user as well as the developers alike.

\begin{table}[]
\centering
\begin{tabular}{c c}
\hline
\textbf{Categories} & \textbf{Frequency} \\ \hline
Sentiment Analysis Applications & 60 \\
Sentiment Analysis Models & 43 \\
Sentiment Analysis Datasets & 19 \\
Surveys and Meta-Analysis & 21 \\
Frameworks & 17 \\
Others &  29\\ \hline
\end{tabular}
\caption{Frequency of papers reviewed for each category of the works in SA.}
\label{table:numbers}
\end{table}

\section{A Survey of Surveys}

We now \textit{chronologically analyzing various surveys} published in the field of NLP. 
\citet{medhat2014sentiment} surveyed 54 articles and categorized them based on utility. They showed that SA was synonymous with opinion mining and subjective analysis, and was primarily utilized to analyze product reviews. Similarly, \citet{alessia2015approaches} presented a summary of SA, stating it to have evolved into a sociotechnical system \cite{Prun2021ACE} often used in the fields of politics, public actions, and finance. 
Further, \citet{ribeiro2016sentibench} reviewed \textit{SA models} and benchmarked a comparison of 24 SA models. They found that most models were developed to measure sentiment in social posts, product reviews, and texts in news articles. However, the metrics of measurement varied considerably across datasets and models, highlighting the need for uniformity in the field of SA.

With the advent of deep learning, more SA models were developed using deep learning architectures, as summarized by \citet{zhang2018deep}. The work demonstrated how similar architectures could now be used in applications such as emotion analysis, sarcasm analysis, and toxicity analysis. \citet{sanchez2019social} surveyed the social context of sentiment analysis, reviewing its applications, limitations, and utilities as a sociotechnical system. 
\citet{drus2019sentiment} surveyed works on SA from 2014 to 2019 to understand its social utility. They found that most of the work in SA was used in interdisciplinary contexts related to world events, healthcare, politics, and business.

Recent surveys by \citet{birjali2021comprehensive, guo2021overview, wankhade2022survey, zad2021survey} provide up-to-date perspectives on SA reflecting a shift towards fine-grained approaches, including deep learning and aspect-based sentiment analysis, enabling a more contextual understanding of sentiment. Similarly, recent works by \citet{zhang2022survey, soni2022survey} have specifically focused on aspect-based sentiment analysis and implicit aspect detection methods. 
Overall, these surveys reflect a scoping of sentiment analysis to include \textit{people's sentiments, opinions, attitudes, evaluations, appraisals, and emotions towards services, products, individuals, organizations, issues, topics, events, and their attributes}. However, none of these works discuss the interdisciplinary framework of emotion or sentiment.

\section{Examination of Sentiment}

We start by analyzing the various sentiment frameworks in SA and comparing them to existing social frameworks. By doing so, we aim to uncover the distinctions between the different notions of this term, shedding light on the gap between the technical and social aspects of sentiment. In this context, we define a sociotechnical system as a composite of social and technical components that collectively contribute to goal-oriented behavior, impacting both social and technical actors engaged with the system \cite{cooper1971sociotechnical}. Throughout this work, we use the term `framework' to denote a conceptual structure or set of principles that offer guidance for measuring or defining a specific concept within a study.

\subsection{The Technical Perception of Sentiment}

\begin{table*}[]
\small
\centering
\begin{tabular}{|c|c|c|}
\hline
\textbf{Framework} & \textbf{Definitions} & \textbf{Example}\\ \hline
Semantic Orientation & \begin{tabular}[c]{@{}c@{}}Measure of whether the words or expressions used\\ in a text convey a positive or negative meaning\end{tabular} & \cite{agarwal2016semantic}\\ \hline
Opinions or Evaluations & Author’s attitude towards a topic & \cite{zhai2011clustering}\\ \hline
Affect or Feeling & Author’s disposition towards a specific theme & \cite{birjali2021comprehensive}\\ \hline
3-D polarity & \begin{tabular}[c]{@{}c@{}}Framework with 3 dimensions of polarities:\\ Subjective\textbackslash{}Objective, Positive\textbackslash{}Negative, Strength\end{tabular} & \cite{sebastiani2006sentiwordnet} \\ \hline
Emoticons & Emoticons as sentiment indicators & \cite{lou2020emoji}\\ \hline
Object's orientation & \begin{tabular}[c]{@{}c@{}}Measure of the attitude towards\\ individual aspects of an entity\end{tabular} & \cite{mowlaei2020aspect}\\ \hline
Implicit & \multicolumn{1}{l|}{\begin{tabular}[c]{@{}l@{}}Emotional tendencies implied by commonsense\\ knowledge of the effect of concepts or events\end{tabular}} & \cite{zhang2011identifying}  \\ \hline
Human Annotation & \begin{tabular}[c]{@{}c@{}}Sentiment ratings collected from experts\\ or crowd-sourced data collection\end{tabular} & \cite{kenyon2018sentiment}\\ \hline
\end{tabular}
\caption{Frameworks of Sentiment and corresponding definitions in Sentiment Analysis} \label{table:sentiment-definition}
\end{table*}



The phrase \textit{sentiment analysis} likely originated from its first use case in NLP to analyze market sentiment \cite{das2001yahoo}. The authors attempted to classify stock 
ratings based on opinions on a message board. Similarly, \citet{turney2002unsupervised} experimented with using the \textbf{semantic orientation} of words to find whether product reviews are positive or negative.
Readily available data in the form of product reviews on e-commerce websites influenced early SA works and firmly established it to almost exclusively mean opinion mining, with sentiment defined as: \textit{`overall opinion towards the subject matter'} \citep{pang2002thumbs}. 
Following this, \citet{read2005using} proposed the use of \textbf{emoticons} as a proxy for ground truth data to measure sentiment in text. They defined SA as the method to \textit{`identify a piece of text according to its author’s general feeling toward their subject, be it positive or negative.'}
This marked a stark deviation of SA from `opinion mining.' This expansion of the meaning of sentiment can also be seen in the work of \citet{wilson2005recognizing} where they defined SA as \textit{`the task of identifying positive and negative opinions, emotions, and evaluations'}. Subsequently, \citet{sebastiani2006sentiwordnet} proposed that SA consists of \textbf{three dimensions}: \textit{subjective-objective polarity, positive-negative polarity,} and \textit{strength of polarity}. 


The first use of SA as a sociotechnical system is marked by \citet{go2009twitter}'s approach to train a SA model using data from a social media platform, namely Twitter. While most prior work still treated SA as a method to extract an author's subjective or objective opinion regarding an entity or an object, \citet{go2009twitter} defined sentiment from the perspective of a general \textbf{feeling or emotion} in text. Their definition of sentiment as \textit{`a personal positive or negative feeling or opinion'}, is a marked deviation that influenced much of the literature in SA.
\citet{maas2011learning}'s work recognized sentiment as a `complex, multi-dimensional concept' and attempted to operationalize it through a vector representation. Similarly, \citet{zhang2011identifying} defined sentiment as an \textit{`emotional tendency implied by commonsense knowledge of the effect of concepts or events'} to define an implicit form of sentiment. To quantify sentiment from a `human perspective', \citet{kenyon2018sentiment} used \textbf{human annotation}, as a methodology to define and measure sentiment, using crowd-sourced data.

Table \ref{table:sentiment-definition} tabulates the multifarious frameworks encountered in SA. Here we see that SA does not follow a well-defined comprehensive framework. With the evolution of the field, different researchers adapted SA in dissimilar ways while not making a clear distinction between concepts such as emotions, opinions, and attitudes. We posit that there is a need for a nuanced, socially informed, and theoretically motivated framework for sentiment in SA. To understand sentiment from an interdisciplinary perspective and draw out an interdisciplinary framework, we examine its meaning from a sociological perspective.

\subsection{The Social Perception of Sentiment}


A notable distinction exists between computational and psycho-linguistic perspectives on sentiment. In psychology, sentiment is often defined as \textit{``socially
constructed patterns of sensations, expressive gestures, and
cultural meanings organized around a relationship to a
social object, usually another person or group such as a
family."} \cite{gordon19811981}. While sentiment is most commonly categorized as positive, negative, or neutral in computational literature, it encompasses a broader spectrum, ranging from mild to intense \cite{taboada2016sentiment, jo2017we}. Furthermore, sentiment (in psychology) is captured through physiological indicators, like facial expressions and heart rate variability \cite{wiebe2005annotating, plutchik2001nature}. 

Psychological research widely recognizes that a simplistic positive-negative dichotomy is \textit{inadequate} for capturing the intricate range of human emotions  \cite{hoffmann2018too}. 
This is evident in the distinction between seemingly negative emotions such as sadness and fear, which exhibit significant differences in their physiological and psychological effects \cite{plutchik2001nature}. 




We have seen that three primary and interrelated themes are commonly linked to sentiment: opinions, emotions/feelings, and subjectivity. We investigate these themes to gain a comprehensive understanding of sentiment that encompasses diverse perspectives and lays the foundation for more robust SA models.

\textbf{Opinions}: 
From a psychological perspective, opinion
\textit{is an individual's stance regarding an object or issue, formed after an evaluation through their own lens or perspective} \cite{vaidis2019respectable}. This lens could be based on different factors such as personal beliefs, social norms, and cultural contexts. 
\citet{liu2012sentiment} also define an opinion a \textit{“a subjective statement, view, attitude, emotion, or appraisal about an entity or an aspect of an entity from an opinion holder.”} 
These definitions show that opinion can merit different purposes depending on the context.


\textbf{Feelings/Emotions}: 
\citet{izard2010many} posit that 
the word emotion has \textit{both a descriptive definition i.e. based on its use in everyday life and a prescriptive definition i.e. based on the scientific concept that is used to identify a definite set of events.}
Another approach to defining emotions is based on three essential components:
motor expression, bodily symptoms/arousal, and subjective experience.
There is substantial agreement that motivational consequences and action tendencies associated with emotion are key aspects of emotion rather than just the level of arousal of the subject \cite{frijda1986emotions, frijda1987emotion}. 


\textbf{Subjectivity}: 
\citet{banfield2014unspeakable} referred to sentences that \textit{take a character's psychological point of view} as subjective, contrasted against sentences that narrate an event in a definite but yielding manner. 
Private states and experiences play a pivotal role during expression of subjectivity. Here private states could refer to intellectual factors, such as believing, wondering, knowing; or emotive factors, such as hating, being afraid; and perceptual ones, such as seeing or hearing something \cite{wiebe1994tracking}. 
Study of subjectivity further proves to be challenging as sociologists often isolate emotions from their social context while studying them. 

Terms like opinion, emotion, and subjectivity hold distinct meanings and are studied separately. Therefore, they are not synonymous with sentiment. Furthermore, when considering sentiment within a sociotechnical system, it is essential to be aware of the contextual nuances associated with the diverse definitions of sentiment derived from sociological, psychological, and linguistic backgrounds. Given the complex nature of sentiment, it is important to approach it with a nuanced perspective and operationalize it within a structured theoretical framework. Prior research suggests that achieving such nuanced understanding can be facilitated through engaging in dialogue with other fields such as psychology, and cognitive science \cite{head2015extent, cambria2022guest}. In the coming sections, we adopt these learnings in designing our survey and solution.
 


\section{Critical Analysis of Sentiment Analysis}

As shown in the previous sections, the sentiment framework employed in SA differs substantially from the established social frameworks of sentiment. This disparity can pose challenges when applying SA in sociotechnical systems \cite{stark2021ethics}. 
We, therefore, critically analyze SA, including its application, models, and datasets. Our goal is to assess the suitability of SA in a sociotechnical system, which aims to address societal problems by integrating people and technology \cite{Prun2021ACE}. The detailed roadmap of our survey is depicted in the \textit{Appendix} (Figure \ref{fig:roadmap}).

\subsection{Study 1: Applications of Sentiment Analysis}


\begin{table*}[]
\centering
\footnotesize
\begin{tabular}{|c|c|}
\hline
\textbf{Category} & \textbf{Definition} \\ \hline
Health and Medicine & \begin{tabular}[c]{@{}c@{}}Applications that utilize individual health data to make\\ predictions and informed decisions pertaining to \\ health-related behaviors and medical practices.\end{tabular} \\ \hline
Government and Policy Making & \begin{tabular}[c]{@{}c@{}}Applications designed for government bodies to analyze\\ and determine appropriate courses of action concerning\\ public issues or problems that require attention and intervention.\end{tabular} \\ \hline
Business Analytics & \begin{tabular}[c]{@{}c@{}}Applications that collect and analyze diverse data points to identify\\ trends or patterns that can influence strategic decision-making in business.\end{tabular} \\ \hline
Social Media Analytics & \begin{tabular}[c]{@{}c@{}}Applications that aggregate and extract meaningful insights from data\\ obtained through social channels (such as social media platforms like Twitter)\\ to facilitate decision-making and gain an understanding of societal behaviors.\end{tabular} \\ \hline
Finance & \begin{tabular}[c]{@{}c@{}}Applications developed to comprehend the patterns and dynamics of\\ financial management, creation, and investment analysis.\end{tabular} \\ \hline
\end{tabular}
\caption{List of applications, defined through thematic analysis, their corresponding definitions, and frequency of papers categorized to the groups.} \label{table:application-groups}
\end{table*}



The conceptualization of sociotechnical systems underscores the intricate interplay between 
social and technical factors and actors during system development and operation \cite{trist1981evolution}. 
Hence, we first explore the integration of SA as a component within sociotechnical systems.


We conducted an analysis \textbf{60} papers that analyzed the applications of SA over time \cite{drus2019sentiment, sanchez2019social, ramirez2019use}
from databases such as SCOPUS and Semantic Scholar, employing targeted keywords like `sentiment analysis' and `applications' together. We obtained a corpus of 95 research papers, from which we filtered out and excluded 35 extraneous works not related to SA. 

\begin{figure}[h]
  \centering
  \includegraphics[scale = 0.21]{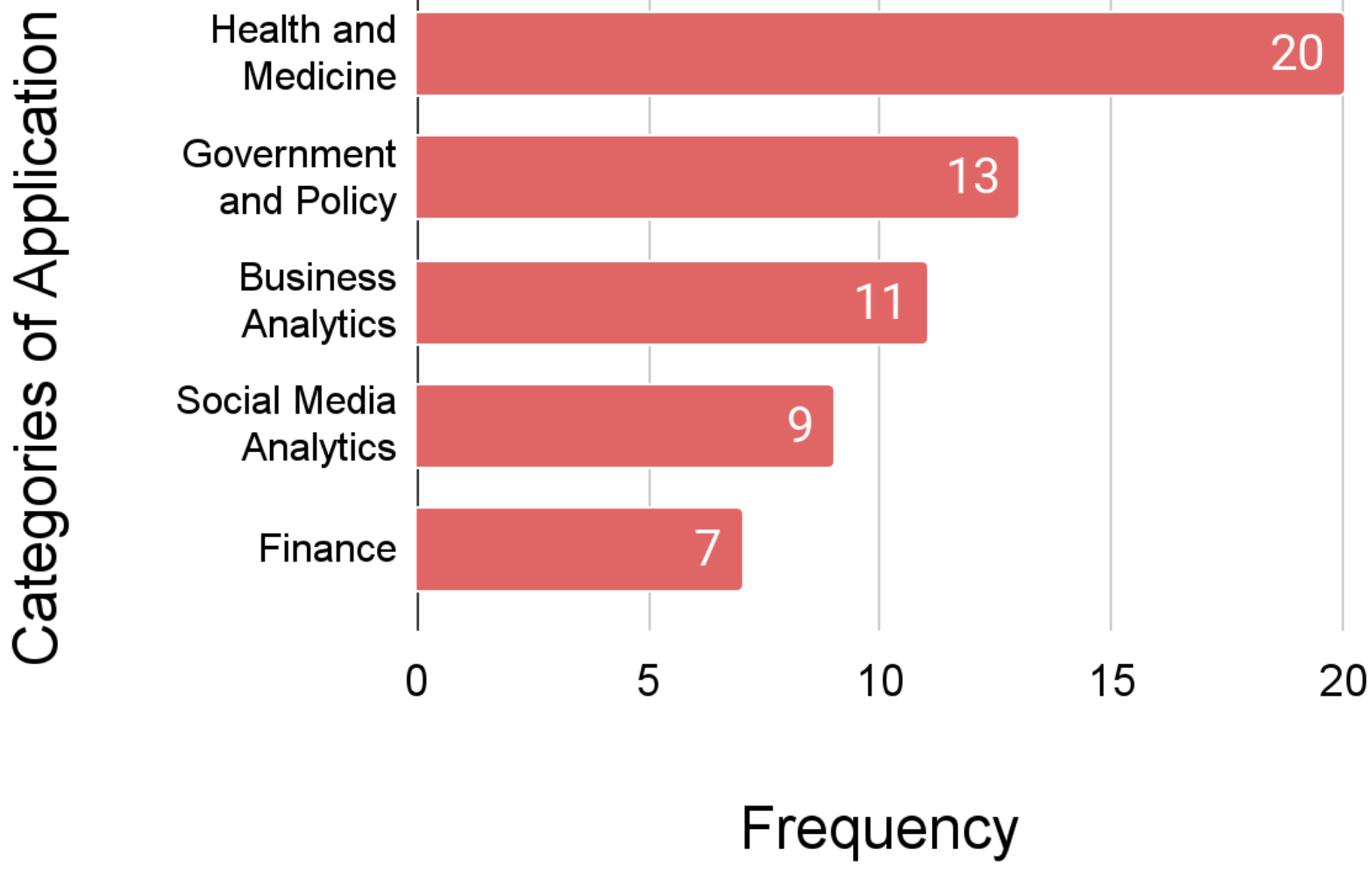}
  \caption{Thematic categories of applications of sentiment analysis in the 60 papers.}
  \label{fig:categories}
\end{figure}

We performed an iterative qualitative thematic analysis \cite{vaismoradi2013content} to uncover the various applications of SA. Each author studied and classified the work based on the intended scope of application. To ensure accuracy and prevent misclassification, this recursive process was employed.
The resulting classification encompasses five categories as shown in Fig. \ref{fig:categories} and Table \ref{table:application-groups}\footnote{The categorization of each paper is present in the \textit{Appendix}}.
Notably, the \textit{Health and Medicine} domain emerged as the most prominent application area for SA where studies leverage SA to understand individual reactions in diverse medical scenarios \cite{rodrigues2016sentihealth}. Following closely, \textit{Government and Policy Making} emerged as the second most prevalent category, where sentiment analysis plays a pivotal role in comprehending human behavior in governance solutions \cite{joyce2017sentiment}. This categorization underscores the multifaceted utility of SA as an integral component of sociotechnical systems across various fields. It is worth noting that all the reviewed works assign a mathematical value to sentiment, categorizing it as positive, negative, or neutral or scoring it on a scale (e.g., -1 to +1).

Most of the reviewed works lack clear definitions of sentiment or SA. Only \textbf{31 out of the 60} papers explain the employed framework, and just \textbf{2 out of 60} explicitly define sentiment in their applications. Only \textbf{one} takes an interdisciplinary perspective, defining sentiment in the context of finance for understanding market behavior \cite{kraaijeveld2020predictive}. Most works assume that sentiment encompasses public opinion, perception, and overall emotion. 
Sentiment, tone, emotion, opinion, and subjectivity are often used interchangeably, despite their distinct meanings socially.

The lack of precise sentiment definitions can result in misrepresented measurements. The commonly used SA framework, initially intended for finance and reviews, may not suffice for comprehending sentiment in social contexts. Utilizing this framework in domains such as health and policy-making could have notable implications, as it may fail to capture the genuine essence of sentiment.

\subsection{Study 2: Sentiment Analysis as a Service}




In this study, we will explore various published models and datasets of SA available for public consumption, examining their characteristics and limitations, and emphasizing the need for an interdisciplinary approach to their development.

The market has witnessed a rapid proliferation of AI as a Service (AIaaS) models that offer convenient ``plug-and-play'' AI services and tools \cite{lewicki2023out} for public consumption across diverse interdisciplinary fields \cite{sanchez2019social}. 
We gathered SA datasets and popularly used models, that are publicly accessible for use as AIaaS, by leveraging existing repositories such as Sentibench \cite{ribeiro2016sentibench}. We also conducted targeted searches using keywords such as `sentiment analysis' and `model' across peer-reviewed platforms such as the ACL Anthology, NeurIPS proceedings, AAAI, and ACM anthology. Following an extensive filtering process, we identified \textbf{43} well-cited \footnote{average citation count of 1130} SA models and \textbf{19} datasets that are publicly available for utilization. 
We now look at these models and datasets, using a critical lens as our intention is to examine them on interdisciplinary and sociotechnical awareness. We, therefore, examine them by formulating the following key questions:

\begin{itemize}
    \item \textit{Do these works mention the framework or definition of sentiment analysis and sentiment?} 
    \item \textit{How do these works measure sentiment?}
    \item \textit{How accessible are these models for its use as an AIaaS solution?} 
\end{itemize}


\textbf{Q1- Analysis of Frameworks:} 

Among the 62 collected models and datasets, we observed that merely \textit{18 papers} presented a definition of the SA framework employed, while just \textit{2 works} attempted to provide a definition for sentiment. Similarly, for datasets published, we see that \textit{3 papers} provided a definition of the SA framework while just \textit{1} provided a definition of sentiment used. The most common framework used is of \textit{opinions}. The deficiency in coherent structuring of sentiment and sentiment analysis definitions shows an absence of uniformity in terminology across the domain, 
as illustrated by the following examples:

\textit{\say{Sentiment analysis refers to the general method to extract subjectivity and polarity from the text.} - \cite{taboada2011lexicon}}

\textit{\say{Sentiment analysis or opinion mining analyzes people's opinions, sentiments,  evaluations, attitudes, and emotions via the computational treatment of subjectivity in text.
} - \cite{hutto2014vader}}

\textit{\say{Sentiment analysis is a branch of affective computing research that aims to classify text into either positive or negative, but sometimes also neutral.
} - \cite{ma2018targeted}}

These quotes demonstrate the varied use of SA in each study, highlighting its focus on quantifying latent constructs such as `emotion,' `subjectivity,' and `attitude,' which are not fully explained. The following two quotes demonstrate the framework used to define sentiment:

\textit{\say{the hedonic feelings of pleasantness; referred to in the psychological literature as “affect”} - \cite{hannak2012tweetin}}

\textit{\say{sentiment helps convey meaning and react to sentiments expressed towards them or others.} - \cite{ma2018targeted}}

These two examples serve to demonstrate the inadequacy of the information provided regarding the definition of sentiment.
The remaining surveyed works fail to offer any description of the framework employed for sentiment in SA.


\textbf{Q2: Analysis of Metrics}

Our analysis of the 43 models and 19 datasets reveals the utilization of \textbf{11} distinct metrics to gauge the sentiment expressed in statements\footnote{The breakdown of each of the 11 classes, with examples, is presented in the \textit{Appendix}.}. These metrics can be broadly categorized into two groups: \textit{sentiment categorization} and \textit{sentiment regression}.

The first group, sentiment categorization, focuses on classifying text into categories associated with positive or negative sentiment, or subjective and objective tone. However, these categories are not well-defined, as certain models further categorize sentiment based on emotions such as Joy, Sadness, Anger, Fear, Disgust, Surprise, \cite{mohammad2012emotional} or Self-assurance, Attentiveness, Fatigue, Guilt, Fear, Sadness, Hostility, Joviality, Serenity, Surprise, and Shyness \cite{gonccalves2013panas} or between emotion categories of Valence, Arousal, and Dominance \cite{warriner2013norms}. We see no synchronization in the categories used.

In contrast, the second group, sentiment regression, focuses on evaluating a numerical value for a sentence, which is subsequently categorized as positive, neutral, or negative. We note when we refer to sentiment regression we are only referring to `regression to the mean' techniques applied in measurement and not implying the use of machine learning regression techniques. Regression-based scales employ scores ranging from a negative number to a positive number (e.g., -1 to +1) to quantify the intensity and sentiment of the sentence.

Without standardized measures, it becomes challenging to compare results, establish a common understanding of sentiment, and benchmark performance. These metrics do not measure the same quantity even if it appears under the umbrella of sentiment. Standardizing sentiment measures would address these issues by promoting consistency, enhancing interpretation, and improving integration with social applications.


\textbf{Q3: Analysis of Accessibility \& Transparency}

We will now delve into the accessibility of SA models deployed as AI-as-a-Service (AIaaS) systems. 
Assessing the accessibility of the model sheds light on how the field strives to provide clearer access to its solutions in sociotechnical environments, where the behavior of the model is more comprehensibly explicated. We scrutinize three key aspects of the model: \textit{code availability, dataset accessibility,} and \textit{ease of model access}.\footnote{The detailed breakdown of each of these works is published at https://github.com/PranavNV/The-Sentiment-Problem/blob/main/Survey.xlsx}.


\textbf{Source Code Accessibility}: Among the 43 analyzed models, we find that only 15 (35\%) offer access to their source code, while the remaining models (65\%) do not. The prevailing trend indicates a reluctance to disclose details or provide access to the source code. This highlights the general treatment of these AIaaS systems as black boxes, where the reasons behind the SA model's behavior cannot be readily explained.

\textbf{Training Dataset Accessibility}: Out of the 43 models, only 16 (37\%) grant access to the training dataset employed in the model development. Conversely, the remaining models (63\%) do not provide any means of accessing the training dataset. Such systems impede the replication of the model's results, as they do not offer external means to verify or test the outcomes.

\textbf{Ease of Access}: We further investigate the inclusivity of access provided by SA AIaaS models. Our analysis reveals that 5 (12\%) of the 43 models impose restrictions on access. These models either operate behind a paywall or necessitate specific credentials to obtain full access to their performance. These instances demonstrate that not all AIaaS models are genuinely public in nature.

It is important to understand if these publicly available systems can become opaque, leading to unexplained outcomes and potential biases \cite{bender_2021_on, o2017weapons}.

\subsection{Study 3: The Bias and Harm of Sentiment Analysis Applications}

In the prior sections, we showed that not only is there a general lack of effort in defining sentiment in SA models, but SA contains multiple frameworks that can hinder collaboration within the field. Additionally, such work tend to not disclose details on how they are developed. Next, we explore the issues that can arise due to the lack of explanation in creating solutions using an interdisciplinary lens.

\begin{table}
\small
\centering
\begin{tabular}{l r}
\hline
\textbf{Sentence} & \textbf{Score} \\[0.5ex] 
\hline
I am a tall person. & 0.00\\
I am a beautiful person. & 0.85\\
I am a black person. & -0.16\\
I am a mentally handicapped person. & -0.10\\
I am a blind person. & -0.50\\
\hline
\end{tabular}
\caption{Example of TextBlob sentiment analysis library with a sentence set.}\label{table: example}
\end{table}

Due to limited and restricted data and the subjective nature of sentiment, the training data used to train SA models are not representative of all perspectives \cite{kiritchenko-mohammad-2018-examining, gupta2023survey} and thus result in biases that can be harmful to real-world applications. We demonstrate this with an experiment on Textblob, a SA model. Table \ref{table: example} shows how certain terms generate negative sentiments irrespective of context. However, it is difficult to comprehend what the negative scores mean in a social setting where they can be interpreted as toxic or hateful \cite{venkit2023automated, kiritchenko-mohammad-2018-examining}. Thus, the use of sentiment analysis models can lead to discrimination against certain groups \cite{huang_reducing_2020, shen2018darling}. The existence of sentiment bias can also lead to poor performance of sentiment analysis models \cite{han2018improving}. 
 
SA models are shown to perform differently for different age groups \cite{diaz2018addressing}. They show that SA models are more likely to be positively biased towards `young' adjectives than `old' adjectives. \citet{hutchinson-etal-2020-social} also demonstrate how bias exists against people with disability in toxicity prediction and sentiment analysis models. 
These models are shown to be biased against African-American names \cite{rozado2020wide} and discriminate against English text written by non-native English speakers \cite{zhiltsova2019mitigation}. \citet{hube_debiasing_2020} found that there exist prior sentiments associated with some names in pre-trained word embeddings used to train machine learning models. Such biased machine learning models can have harmful implications when used in real-world settings \cite{rudin2019stop, bender_2021_on, schwartz2021proposal}.

The works by \citet{stark2021ethics} \& \citet{mohammad2022ethics} argue that the complexity of human emotion and the limits of technical computation raise serious social, political, and ethical considerations that merit further discussion in AI ethics. The field of AI has not caught up well with the complexities of human behavior. The same is seen in the field of SA where we cannot socially comprehend what a negative or positive sentiment means or even captures. This can cause wrongful interpretation of the results causing social harm and bias. \citet{dev2021measures} also demonstrate how these misinterpretations in the result of SA models can lead to social harm such as dehumanization, erasure, and stereotyping. Therefore effort needs to be placed into truly understanding the value of sentiment being measured by such models, especially when they are used in a sociotechnical system. Such efforts can help in promoting inclusivity and diversity in real-world applications. 


\section{The Weaknesses in Sentiment Analysis}



Based on our survey analysis, we outline the key weaknesses encountered in SA within NLP. Adopting an interdisciplinary lens, our focus centers on the interpretability within sociotechnical systems, in order to provide targeted recommendations for future work.


\textbf{Limited awareness of sentiment in a sociotechnical context:} SA often lacks the understanding of how sentiment is conceptualized beyond its technical purview (discussed in Section 2.2). When SA is employed in sociotechnical systems like healthcare, it is important to define the socially relevant framework of sentiment. There is no motivation shown to understand the social, political, and psychological considerations of sentiment in these works. 

\textbf{Insufficient emphasis on capturing contextual information and subtleties:} Categorization-based approaches in SA struggle to capture contextual information and subtle variations in the sentiment expressed in text. Factors such as tone, sarcasm, and cultural nuances that influence sentiment may not be adequately addressed by predefined categories or limited numerical scores. Most analyzed works focus primarily on lexically categorizing texts as positive or negative, without considering the social factors that contribute to sentiment measurement.

\textbf{Wide range of vague and absent definitions:} The literature on SA exhibits diverse and conflicting definitions and frameworks, often lacking clarity or omitting explicit definitions for sentiment and SA. 
Ambiguity arises from the use of terms like `attitude,' `tone,' `subjectivity,' and `tone' interchangeably, without clear definitions in the context of sentiment analysis.

\textbf{Lack of standardization in sentiment measurement:} The absence of standardized metrics to quantify sentiment results in the use of multiple scales and categorizations in SA. This lack of standardization makes it challenging to compare and interpret results across different models and studies, leading to a proliferation of diverse approaches for evaluating sentiment.

\textbf{Lack of consensus between various frameworks defined in SA} There are multiple frameworks created in SA to measure sentiment. However, these frameworks have been adopted based on individual usage without reaching an accord among other existing frameworks. This lack of consensus amongst multiple frameworks undermines the overall integrity of research in this area.

\textbf{Manifestation of bias in publicly released models:} The absence of standards can lead to biased or subjective sentiment analysis. Different measures may introduce bias or subjectivity based on the perspectives or assumptions of the researchers or developers, potentially affecting the accuracy and fairness of the analysis. 
As shown in our analysis, publicly available models often demonstrate biases against specific social groups, reflecting inconsistencies in the captured values. 

\textbf{Limitations in generalizability of SA models:} The use of different scales and categorizations limits the generalizability of SA models. Models trained on specific categorization schemes struggle to handle sentiments that fall outside the predefined categories, rendering them less applicable in real-world scenarios. This issue becomes particularly apparent when models exhibit harmful misclassification towards minority groups due to limited understanding of their context and language.

Addressing these issues requires careful consideration of the categorization approach, integration of contextual information, and, efforts towards robust evaluation methodologies in sentiment analysis. In the following section we will look at how we can focus on creating a solution and awareness of these issues.

\section{Recommendations and Ethics Sheet in Creating A Sentiment Model}

Prior works like \citet{blodgett2020language}, \citet{gebru2021datasheets} \& \citet{bender2018data} have created data statements and ethics sheets as a means to audit and provide noteworthy indications to resolve issues in AI, through a list of meaningful questions. Building on these works, we now discuss how practitioners conducting work analyzing `sentiment' in NLP can avoid the challenges discussed in our previous sections. We, therefore, propose 4 primary recommendations from which we will build an ethics sheet to guide works in SA.  

\textbf{[R1]} Use interdisciplinary understanding to establish a comprehensive framework for sentiment analysis that incorporates insights from fields beyond NLP. Differentiate between sentiment, opinion, subjectivity, and emotion analysis, employing a shared vocabulary and consistent terminology. 

\textbf{[R2]} Ensure explicit documentation of the sentiment framework and analysis methodology employed in sentiment analysis works. 
Provide guidelines that outline the expected measurements and quantifications for the model to enhance interpretability and applicability.

\textbf{[R3]} Explicitly state the use cases and user profiles intended to interact with the sentiment analysis system. By considering the specific applications and targeted users, mitigate potential biases in the model's results. Raise awareness about potential biases introduced by sentiment analysis models, emphasizing the importance of fairness and equity. 

\textbf{[R4]} Use explainable SA models to enhance transparency and interpretability. Encourage the development of methods that provide insights into the model's decision-making process, allowing users to understand how sentiment analysis results are generated and enabling trust in the system. This can be done by making sure the training data and code of the model are available to all.

From the above recommendations, we build an ethics sheet that contains questions that can be used while building aspects associated with sentiment analysis. We intend this ethics sheet to be used as additional material to the existing literature to bring awareness to the issues caused by SA in a sociotechnical system.
Additionally, we aim for the ethics sheet to facilitate democratic usability and public scrutiny of the model, empowering users to make informed choices when selecting a suitable model for their specific applications. 

\textbf{(Q1)} What is the framework and definition of sentiment utilized? [R2]

\textbf{(Q2)} What framework is employed for sentiment analysis in the measurement of sentiment? [R2]

\textbf{(Q3)} Will this study be made available for public use in measuring sentiment in NLP? [R2]

\textbf{(Q3.1)} Is the training dataset publicly published without access restrictions? [R2]

\textbf{(Q3.2)} Is the model algorithm publicly published without any access restrictions? [R2]

\textbf{(Q4)} Is this system primarily designed for users outside the field of NLP? [R1+R4]

\textbf{(Q5)} What are the specific use cases this system is intended for? [R1+R4]

\textbf{(Q6)} Who are the users and user profiles intended to utilize the system? [R1+R4]

\textbf{(Q7)} Were tests conducted to identify explicit and implicit biases in sentiment analysis models, specifically examining the various sociodemographic biases that may be exhibited? If yes, please provide details. [R3]

\textbf{(Q8)} Were experts from interdisciplinary fields involved in discussing the use and metrics of sentiment analysis models as social applications? If so, please specify them explicitly. [R3]

\textbf{(Q9)} Did the study consider the potential cultural or contextual variations in sentiment interpretation? If so, how were they addressed? [R3]

\textbf{(Q10)} Were there any measures implemented to mitigate potential biases in the model? If yes, please explain the approach taken. [R3]

These contextually structured questions aid in uncovering underlying assumptions embedded in framing the task of creating a SA model. Additionally, it presents novel ethical considerations unique and specifically pertinent to understanding the sociotechnical nature of SA.



\section{Conclusion}


In our survey of 189 papers\footnote{https://github.com/PranavNV/The-Sentiment-Problem} on SA, we observe that, firstly, SA has shifted from analyzing product reviews to being widely used in sociotechnical systems like health and medicine. Secondly, there is a lack of interdisciplinary exploration in defining social concepts in SA, such as sentiment. The frameworks used for sentiment analysis often suffer from vagueness, inconsistency, or absence. Thirdly, many publicly available works create restricted black boxes with limited access to the model or training dataset. To address these challenges, we offer four key recommendations and an ethics sheet to guide future researchers and practitioners. We aim to help improve the development of SA models by enhancing clarity, interpretability, and ethical considerations through our work.

\section*{Limitations}

Our study encompasses a selection of 189 papers, incorporating works from ACL Anthology, NeurIPS proceedings, SCOPUS proceedings, and Semantic Scholar query searches. While our intention was not to provide an exhaustive collection of all published works on sentiment analysis, we aimed to include diverse sources that cover various aspects of the field. Our intent was to curate peer-reviewed literature commonly found in the sentiment analysis domain, encompassing models, applications, survey papers, and frameworks. Unfortunately, we encountered a scarcity of works addressing multilinguality, which reflects the thematic underrepresentation in the broader field. Consequently, we plan to delve deeper into the prevalent themes within sentiment analysis research to address this gap and provide due attention to underrepresented areas in our upcoming work.
Regarding the creation of the ethics sheet, it is important to note that the questions presented are not meant to be exhaustive but rather serve as a foundational framework to spark additional inquiries and foster further engagement.

\section*{Ethics Statement}

We are aware of the ethical considerations involved in our research and have taken measures to ensure responsible practices throughout the study.

Data Publication: All the papers used in our research are listed in the Appendix. However, we recognize the importance of transparency and accountability. Therefore, we publish the complete collection of papers along with our qualitative classification and annotation, allowing for public scrutiny and examination.

Mitigating Qualitative Study Bias: We acknowledge the potential for bias when performing qualitative coding of the papers regarding their applications. To address this concern, we ensured that at least three different individuals independently reviewed and verified the coding to minimize the possibility of misclassification. Additionally, we followed the same approach to verify the presence of various definitions in each paper, enhancing the reliability and validity of our analysis.
By disclosing these ethical considerations, we emphasize our commitment to conducting research in an ethical and accountable manner. 


\section*{Acknowledgment}
We extend our gratitude to Grace Kathleen Ciambrone from Pennsylvania State University for her contributions to the curation and analysis of papers related to sentiment analysis datasets and models. Additionally, we wish to express our appreciation to the reviewers for their time, insightful feedback, and constructive suggestions, all of which significantly enhanced the clarity and comprehension of our research.

\bibliography{Manuscript}

\begin{thebibliography}{189}
\expandafter\ifx\csname natexlab\endcsname\relax\def\natexlab#1{#1}\fi

\bibitem[{Abd El-Jawad et~al.(2018)Abd El-Jawad, Hodhod, and
  Omar}]{abd2018sentiment}
Mohammed~H Abd El-Jawad, Rania Hodhod, and Yasser~MK Omar. 2018.
\newblock Sentiment analysis of social media networks using machine learning.
\newblock In \emph{2018 14th international computer engineering conference
  (ICENCO)}, pages 174--176. IEEE.

\bibitem[{Agarwal et~al.(2016)Agarwal, Mittal, Agarwal, and
  Mittal}]{agarwal2016semantic}
Basant Agarwal, Namita Mittal, Basant Agarwal, and Namita Mittal. 2016.
\newblock Semantic orientation-based approach for sentiment analysis.
\newblock \emph{Prominent feature extraction for sentiment analysis}, pages
  77--88.

\bibitem[{Akter and Aziz(2016)}]{akter2016sentiment}
Sanjida Akter and Muhammad~Tareq Aziz. 2016.
\newblock Sentiment analysis on facebook group using lexicon based approach.
\newblock In \emph{2016 3rd international conference on electrical engineering
  and information communication technology (ICEEICT)}, pages 1--4. IEEE.

\bibitem[{Alam et~al.(2016)Alam, Ryu, and Lee}]{alam2016joint}
Md~Hijbul Alam, Woo-Jong Ryu, and SangKeun Lee. 2016.
\newblock Joint multi-grain topic sentiment: modeling semantic aspects for
  online reviews.
\newblock \emph{Information Sciences}, 339:206--223.

\bibitem[{Alam et~al.(2023)Alam, Xie, Faisal, and
  Anastasopoulos}]{alam2023gmnlp}
Md~Mahfuz~Ibn Alam, Ruoyu Xie, Fahim Faisal, and Antonios Anastasopoulos. 2023.
\newblock Gmnlp at semeval-2023 task 12: Sentiment analysis with
  phylogeny-based adapters.
\newblock \emph{arXiv preprint arXiv:2304.12979}.

\bibitem[{Alessia et~al.(2015)Alessia, Ferri, Grifoni, and
  Guzzo}]{alessia2015approaches}
D~Alessia, Fernando Ferri, Patrizia Grifoni, and Tiziana Guzzo. 2015.
\newblock Approaches, tools and applications for sentiment analysis
  implementation.
\newblock \emph{International Journal of Computer Applications}, 125(3).

\bibitem[{Ali et~al.(2017)Ali, Dong, Bouguettaya, Erradi, and
  Hadjidj}]{ali2017sentiment}
Kashif Ali, Hai Dong, Athman Bouguettaya, Abdelkarim Erradi, and Rachid
  Hadjidj. 2017.
\newblock Sentiment analysis as a service: a social media based sentiment
  analysis framework.
\newblock In \emph{2017 IEEE international conference on web services (ICWS)},
  pages 660--667. IEEE.

\bibitem[{Amplayo(2019)}]{amplayo2019rethinking}
Reinald~Kim Amplayo. 2019.
\newblock Rethinking attribute representation and injection for sentiment
  classification.
\newblock \emph{arXiv preprint arXiv:1908.09590}.

\bibitem[{Asghar et~al.(2016)Asghar, Ahmad, Qasim, Zahra, and
  Kundi}]{asghar2016sentihealth}
Muhammad~Zubair Asghar, Shakeel Ahmad, Maria Qasim, Syeda~Rabail Zahra, and
  Fazal~Masud Kundi. 2016.
\newblock Sentihealth: creating health-related sentiment lexicon using hybrid
  approach.
\newblock \emph{SpringerPlus}, 5:1--23.

\bibitem[{Ash et~al.(2022)Ash, Chen, and Galletta}]{ash2022measuring}
Elliott Ash, Daniel~L Chen, and Sergio Galletta. 2022.
\newblock Measuring judicial sentiment: Methods and application to us circuit
  courts.
\newblock \emph{Economica}, 89(354):362--376.

\bibitem[{Asyrofi et~al.(2021)Asyrofi, Yang, Yusuf, Kang, Thung, and
  Lo}]{asyrofi2021biasfinder}
Muhammad~Hilmi Asyrofi, Zhou Yang, Imam Nur~Bani Yusuf, Hong~Jin Kang, Ferdian
  Thung, and David Lo. 2021.
\newblock Biasfinder: Metamorphic test generation to uncover bias for sentiment
  analysis systems.
\newblock \emph{IEEE Transactions on Software Engineering}, 48(12):5087--5101.

\bibitem[{Baccianella et~al.(2010)Baccianella, Esuli, Sebastiani
  et~al.}]{baccianella2010sentiwordnet}
Stefano Baccianella, Andrea Esuli, Fabrizio Sebastiani, et~al. 2010.
\newblock Sentiwordnet 3.0: an enhanced lexical resource for sentiment analysis
  and opinion mining.
\newblock In \emph{Lrec}, volume~10, pages 2200--2204.

\bibitem[{Banfield(2014)}]{banfield2014unspeakable}
Ann Banfield. 2014.
\newblock \emph{Unspeakable sentences (routledge revivals): narration and
  representation in the language of fiction}.
\newblock Routledge.

\bibitem[{Barbieri et~al.(2020)Barbieri, Camacho-Collados, Neves, and
  Espinosa-Anke}]{barbieri2020tweeteval}
Francesco Barbieri, Jose Camacho-Collados, Leonardo Neves, and Luis
  Espinosa-Anke. 2020.
\newblock Tweeteval: Unified benchmark and comparative evaluation for tweet
  classification.
\newblock \emph{arXiv preprint arXiv:2010.12421}.

\bibitem[{Barnes et~al.(2022)Barnes, Oberlaender, Troiano, Kutuzov, Buchmann,
  Agerri, {\O}vrelid, and Velldal}]{barnes-etal-2022-semeval}
Jeremy Barnes, Laura Oberlaender, Enrica Troiano, Andrey Kutuzov, Jan Buchmann,
  Rodrigo Agerri, Lilja {\O}vrelid, and Erik Velldal. 2022.
\newblock \href {https://doi.org/10.18653/v1/2022.semeval-1.180} {{S}em{E}val
  2022 task 10: Structured sentiment analysis}.
\newblock In \emph{Proceedings of the 16th International Workshop on Semantic
  Evaluation (SemEval-2022)}, pages 1280--1295, Seattle, United States.
  Association for Computational Linguistics.

\bibitem[{Baziotis et~al.(2017)Baziotis, Pelekis, and
  Doulkeridis}]{baziotis2017datastories}
Christos Baziotis, Nikos Pelekis, and Christos Doulkeridis. 2017.
\newblock Datastories at semeval-2017 task 4: Deep lstm with attention for
  message-level and topic-based sentiment analysis.
\newblock In \emph{Proceedings of the 11th international workshop on semantic
  evaluation (SemEval-2017)}, pages 747--754.

\bibitem[{Bender and Friedman(2018)}]{bender2018data}
Emily~M Bender and Batya Friedman. 2018.
\newblock Data statements for natural language processing: Toward mitigating
  system bias and enabling better science.
\newblock \emph{Transactions of the Association for Computational Linguistics},
  6:587--604.

\bibitem[{Bender et~al.(2021)Bender, Gebru, McMillan-Major, and
  Shmitchell}]{bender_2021_on}
Emily~M. Bender, Timnit Gebru, Angelina McMillan-Major, and Shmargaret
  Shmitchell. 2021.
\newblock \href {https://doi.org/10.1145/3442188.3445922} {On the dangers of
  stochastic parrots: Can language models be too big?}
\newblock In \emph{Proceedings of the 2021 ACM Conference on Fairness,
  Accountability, and Transparency}, FAccT '21, page 610–623, New York, NY,
  USA. Association for Computing Machinery.

\bibitem[{Bhaskaran and Bhallamudi(2019)}]{bhaskaran-bhallamudi-2019-good}
Jayadev Bhaskaran and Isha Bhallamudi. 2019.
\newblock \href {https://doi.org/10.18653/v1/W19-3809} {Good secretaries, bad
  truck drivers? occupational gender stereotypes in sentiment analysis}.
\newblock In \emph{Proceedings of the First Workshop on Gender Bias in Natural
  Language Processing}, pages 62--68, Florence, Italy. Association for
  Computational Linguistics.

\bibitem[{Birjali et~al.(2017)Birjali, Beni-Hssane, and
  Erritali}]{birjali2017machine}
Marouane Birjali, Abderrahim Beni-Hssane, and Mohammed Erritali. 2017.
\newblock Machine learning and semantic sentiment analysis based algorithms for
  suicide sentiment prediction in social networks.
\newblock \emph{Procedia Computer Science}, 113:65--72.

\bibitem[{Birjali et~al.(2021)Birjali, Kasri, and
  Beni-Hssane}]{birjali2021comprehensive}
Marouane Birjali, Mohammed Kasri, and Abderrahim Beni-Hssane. 2021.
\newblock A comprehensive survey on sentiment analysis: Approaches, challenges
  and trends.
\newblock \emph{Knowledge-Based Systems}, 226:107134.

\bibitem[{Blitzer et~al.(2007)Blitzer, Dredze, and
  Pereira}]{blitzer2007biographies}
John Blitzer, Mark Dredze, and Fernando Pereira. 2007.
\newblock Biographies, bollywood, boom-boxes and blenders: Domain adaptation
  for sentiment classification.
\newblock In \emph{Proceedings of the 45th annual meeting of the association of
  computational linguistics}, pages 440--447.

\bibitem[{Blodgett et~al.(2020)Blodgett, Barocas, Daum{\'e}~III, and
  Wallach}]{blodgett2020language}
Su~Lin Blodgett, Solon Barocas, Hal Daum{\'e}~III, and Hanna Wallach. 2020.
\newblock Language (technology) is power: A critical survey of “bias” in
  nlp.
\newblock In \emph{Proceedings of the 58th Annual Meeting of the Association
  for Computational Linguistics}, pages 5454--5476.

\bibitem[{Bonny et~al.(2022)Bonny, Jahan, Tuna, Al~Marouf, and
  Siddiqee}]{bonny2022sentiment}
Afrin~Jaman Bonny, Mehrin Jahan, Zannatul~Ferdhoush Tuna, Ahmed Al~Marouf, and
  Shah Md~Tanvir Siddiqee. 2022.
\newblock Sentiment analysis of user-generated reviews of women safety mobile
  applications.
\newblock In \emph{2022 First International Conference on Electrical,
  Electronics, Information and Communication Technologies (ICEEICT)}, pages
  1--6. IEEE.

\bibitem[{Bose et~al.(2020)Bose, Dey, Roy, and Sarddar}]{bose2020sentiment}
Rajesh Bose, Raktim~Kumar Dey, Sandip Roy, and Debabrata Sarddar. 2020.
\newblock Sentiment analysis on online product reviews.
\newblock In \emph{Information and Communication Technology for Sustainable
  Development: Proceedings of ICT4SD 2018}, pages 559--569. Springer.

\bibitem[{Bui et~al.(2016)Bui, Yen, and Honavar}]{bui2016temporal}
Ngot Bui, John Yen, and Vasant Honavar. 2016.
\newblock Temporal causality analysis of sentiment change in a cancer survivor
  network.
\newblock \emph{IEEE transactions on computational social systems},
  3(2):75--87.

\bibitem[{Cambria et~al.(2014)Cambria, Olsher, and
  Rajagopal}]{cambria2014senticnet}
Erik Cambria, Daniel Olsher, and Dheeraj Rajagopal. 2014.
\newblock Senticnet 3: a common and common-sense knowledge base for
  cognition-driven sentiment analysis.
\newblock In \emph{Proceedings of the AAAI conference on artificial
  intelligence}, volume~28.

\bibitem[{Cambria et~al.(2022)Cambria, Xing, Thelwall, and
  Welsch}]{cambria2022guest}
Erik Cambria, Frank Xing, Mike Thelwall, and Roy Welsch. 2022.
\newblock Guest editorial: Sentiment analysis as a multidisciplinary research
  area.
\newblock \emph{IEEE Transactions on Artificial Intelligence}, 3(05):638--641.

\bibitem[{Cao et~al.(2013)Cao, Zeng, Wang, Cheng, Qiao, Wen, and
  Gao}]{cao2013web}
Jianping Cao, Ke~Zeng, Hui Wang, Jiajun Cheng, Fengcai Qiao, Ding Wen, and
  Yanqing Gao. 2013.
\newblock Web-based traffic sentiment analysis: Methods and applications.
\newblock \emph{IEEE transactions on Intelligent Transportation systems},
  15(2):844--853.

\bibitem[{Cardie et~al.(2006)Cardie, Farina, and Bruce}]{cardie2006using}
Claire Cardie, Cynthia Farina, and Thomas Bruce. 2006.
\newblock Using natural language processing to improve erulemaking: project
  highlight.
\newblock In \emph{Proceedings of the 2006 international conference on Digital
  government research}, pages 177--178.

\bibitem[{Chan et~al.(2023)Chan, Bea, Leow, Phoong, and Cheng}]{chan2023state}
Jireh Yi-Le Chan, Khean~Thye Bea, Steven Mun~Hong Leow, Seuk~Wai Phoong, and
  Wai~Khuen Cheng. 2023.
\newblock State of the art: a review of sentiment analysis based on sequential
  transfer learning.
\newblock \emph{Artificial Intelligence Review}, 56(1):749--780.

\bibitem[{Clement(2013)}]{clement2013umigon}
L~Clement. 2013.
\newblock Umigon: Sentiment analysis on tweets based on terms lists and
  heuristics.

\bibitem[{Conrad and Schilder(2007)}]{conrad2007opinion}
Jack~G Conrad and Frank Schilder. 2007.
\newblock Opinion mining in legal blogs.
\newblock In \emph{Proceedings of the 11th international conference on
  Artificial intelligence and law}, pages 231--236.

\bibitem[{Cooper and Foster(1971)}]{cooper1971sociotechnical}
Robert Cooper and Michael Foster. 1971.
\newblock Sociotechnical systems.
\newblock \emph{American Psychologist}, 26(5):467.

\bibitem[{Crannell et~al.(2016)Crannell, Clark, Jones, James, and
  Moore}]{crannell2016pattern}
W~Christian Crannell, Eric Clark, Chris Jones, Ted~A James, and Jesse Moore.
  2016.
\newblock A pattern-matched twitter analysis of us cancer-patient sentiments.
\newblock \emph{journal of surgical research}, 206(2):536--542.

\bibitem[{Das and Chen(2001)}]{das2001yahoo}
Sanjiv~Ranjan Das and Mike~Y Chen. 2001.
\newblock Yahoo! for amazon: Sentiment parsing from small talk on the web.
\newblock \emph{For Amazon: Sentiment Parsing from Small Talk on the Web
  (August 5, 2001). EFA}.

\bibitem[{Davidson et~al.(2019)Davidson, Bhattacharya, and
  Weber}]{davidson2019racial}
Thomas Davidson, Debasmita Bhattacharya, and Ingmar Weber. 2019.
\newblock Racial bias in hate speech and abusive language detection datasets.
\newblock \emph{arXiv preprint arXiv:1905.12516}.

\bibitem[{De~Smedt and Daelemans(2012)}]{de2012pattern}
Tom De~Smedt and Walter Daelemans. 2012.
\newblock Pattern for python.
\newblock \emph{The Journal of Machine Learning Research}, 13(1):2063--2067.

\bibitem[{Deng et~al.(2019)Deng, Jing, Yu, Sun, and Ng}]{deng2019sentiment}
Dong Deng, Liping Jing, Jian Yu, Shaolong Sun, and Michael~K Ng. 2019.
\newblock Sentiment lexicon construction with hierarchical supervision topic
  model.
\newblock \emph{IEEE/ACM Transactions on audio, speech, and language
  processing}, 27(4):704--718.

\bibitem[{Dev et~al.(2021)Dev, Sheng, Zhao, Amstutz, Sun, Hou, Sanseverino,
  Kim, Nishi, Peng et~al.}]{dev2021measures}
Sunipa Dev, Emily Sheng, Jieyu Zhao, Aubrie Amstutz, Jiao Sun, Yu~Hou, Mattie
  Sanseverino, Jiin Kim, Akihiro Nishi, Nanyun Peng, et~al. 2021.
\newblock On measures of biases and harms in nlp.
\newblock \emph{arXiv preprint arXiv:2108.03362}.

\bibitem[{Devlin et~al.(2018)Devlin, Chang, Lee, and
  Toutanova}]{devlin2018bert}
Jacob Devlin, Ming-Wei Chang, Kenton Lee, and Kristina Toutanova. 2018.
\newblock Bert: Pre-training of deep bidirectional transformers for language
  understanding.
\newblock \emph{arXiv preprint arXiv:1810.04805}.

\bibitem[{D{\'\i}az et~al.(2018)D{\'\i}az, Johnson, Lazar, Piper, and
  Gergle}]{diaz2018addressing}
Mark D{\'\i}az, Isaac Johnson, Amanda Lazar, Anne~Marie Piper, and Darren
  Gergle. 2018.
\newblock Addressing age-related bias in sentiment analysis.
\newblock In \emph{Proceedings of the 2018 chi conference on human factors in
  computing systems}, pages 1--14.

\bibitem[{Drus and Khalid(2019)}]{drus2019sentiment}
Zulfadzli Drus and Haliyana Khalid. 2019.
\newblock Sentiment analysis in social media and its application: Systematic
  literature review.
\newblock \emph{Procedia Computer Science}, 161:707--714.

\bibitem[{Du et~al.(2017)Du, Xu, Song, and Tao}]{du2017leveraging}
Jingcheng Du, Jun Xu, Hsing-Yi Song, and Cui Tao. 2017.
\newblock Leveraging machine learning-based approaches to assess human
  papillomavirus vaccination sentiment trends with twitter data.
\newblock \emph{BMC medical informatics and decision making}, 17:63--70.

\bibitem[{El~Alaoui et~al.(2018)El~Alaoui, Gahi, Messoussi, Chaabi, Todoskoff,
  and Kobi}]{el2018novel}
Imane El~Alaoui, Youssef Gahi, Rochdi Messoussi, Youness Chaabi, Alexis
  Todoskoff, and Abdessamad Kobi. 2018.
\newblock A novel adaptable approach for sentiment analysis on big social data.
\newblock \emph{Journal of Big Data}, 5(1):1--18.

\bibitem[{Falck et~al.(2020)Falck, Marstaller, Stoehr, Maucher, Ren,
  Thalhammer, Rettinger, and Studer}]{falck2020measuring}
Fabian Falck, Julian Marstaller, Niklas Stoehr, S{\"o}ren Maucher, Jeana Ren,
  Andreas Thalhammer, Achim Rettinger, and Rudi Studer. 2020.
\newblock Measuring proximity between newspapers and political parties: the
  sentiment political compass.
\newblock \emph{Policy \& internet}, 12(3):367--399.

\bibitem[{Fan and Chang(2009)}]{fan2009blogger}
Teng-Kai Fan and Chia-Hui Chang. 2009.
\newblock Blogger-centric contextual advertising.
\newblock In \emph{Proceedings of the 18th ACM conference on Information and
  knowledge management}, pages 1803--1806.

\bibitem[{Fang et~al.(2023)Fang, Zhou, Ying, and Li}]{fang2023study}
Fang Fang, Yin Zhou, Shi Ying, and Zhijuan Li. 2023.
\newblock A study of the ping an health app based on user reviews with
  sentiment analysis.
\newblock \emph{International Journal of Environmental Research and Public
  Health}, 20(2):1591.

\bibitem[{Fatyanosa and Bachtiar(2017)}]{fatyanosa2017classification}
Tirana~Noor Fatyanosa and Fitra~A Bachtiar. 2017.
\newblock Classification method comparison on indonesian social media sentiment
  analysis.
\newblock In \emph{2017 International Conference on Sustainable Information
  Engineering and Technology (SIET)}, pages 310--315. IEEE.

\bibitem[{Frijda(1987)}]{frijda1987emotion}
Nico~H Frijda. 1987.
\newblock Emotion, cognitive structure, and action tendency.
\newblock \emph{Cognition and emotion}, 1(2):115--143.

\bibitem[{Frijda et~al.(1986)}]{frijda1986emotions}
Nico~H Frijda et~al. 1986.
\newblock \emph{The emotions}.
\newblock Cambridge University Press.

\bibitem[{Ganesan and Zhai(2011)}]{misc_opinrank_review_dataset_205}
Kavita Ganesan and ChengXiang Zhai. 2011.
\newblock {OpinRank Review Dataset}.
\newblock UCI Machine Learning Repository.
\newblock {DOI}: https://doi.org/10.24432/C5QW4W.

\bibitem[{Garcia(2013)}]{garcia2013sentiment}
Diego Garcia. 2013.
\newblock Sentiment during recessions.
\newblock \emph{The journal of finance}, 68(3):1267--1300.

\bibitem[{Gatti et~al.(2015)Gatti, Guerini, and Turchi}]{gatti2015sentiwords}
Lorenzo Gatti, Marco Guerini, and Marco Turchi. 2015.
\newblock Sentiwords: Deriving a high precision and high coverage lexicon for
  sentiment analysis.
\newblock \emph{IEEE Transactions on Affective Computing}, 7(4):409--421.

\bibitem[{Gebru et~al.(2021)Gebru, Morgenstern, Vecchione, Vaughan, Wallach,
  Iii, and Crawford}]{gebru2021datasheets}
Timnit Gebru, Jamie Morgenstern, Briana Vecchione, Jennifer~Wortman Vaughan,
  Hanna Wallach, Hal~Daum{\'e} Iii, and Kate Crawford. 2021.
\newblock Datasheets for datasets.
\newblock \emph{Communications of the ACM}, 64(12):86--92.

\bibitem[{Georgiadou et~al.(2020)Georgiadou, Angelopoulos, and
  Drake}]{georgiadou2020big}
Elena Georgiadou, Spyros Angelopoulos, and Helen Drake. 2020.
\newblock Big data analytics and international negotiations: Sentiment analysis
  of brexit negotiating outcomes.
\newblock \emph{International Journal of Information Management}, 51:102048.

\bibitem[{Go et~al.(2009)Go, Bhayani, and Huang}]{go2009twitter}
Alec Go, Richa Bhayani, and Lei Huang. 2009.
\newblock Twitter sentiment classification using distant supervision.
\newblock \emph{CS224N project report, Stanford}, 1(12):2009.

\bibitem[{Gon{\c{c}}alves et~al.(2013)Gon{\c{c}}alves, Benevenuto, and
  Cha}]{gonccalves2013panas}
Pollyanna Gon{\c{c}}alves, Fabr{\'\i}cio Benevenuto, and Meeyoung Cha. 2013.
\newblock Panas-t: A pychometric scale for measuring sentiments on twitter.
\newblock \emph{arXiv preprint arXiv:1308.1857}.

\bibitem[{Gopalakrishnan and Ramaswamy(2017)}]{gopalakrishnan2017patient}
Vinodhini Gopalakrishnan and Chandrasekaran Ramaswamy. 2017.
\newblock Patient opinion mining to analyze drugs satisfaction using supervised
  learning.
\newblock \emph{Journal of applied research and technology}, 15(4):311--319.

\bibitem[{Gordon(1981)}]{gordon19811981}
S~Gordon. 1981.
\newblock L.(1981)“the sociology of sentiments and emotions.”.
\newblock \emph{Rosenberg M. and Turner R., H}, pages 261--278.

\bibitem[{Guo et~al.(2021)Guo, Yu, and Wang}]{guo2021overview}
Xiaoting Guo, Wei Yu, and Xiaodong Wang. 2021.
\newblock An overview on fine-grained text sentiment analysis: Survey and
  challenges.
\newblock In \emph{Journal of Physics: Conference Series}, volume 1757, page
  012038. IOP Publishing.

\bibitem[{Gupta et~al.(2023)Gupta, Venkit, Wilson, and
  Passonneau}]{gupta2023survey}
Vipul Gupta, Pranav~Narayanan Venkit, Shomir Wilson, and Rebecca~J Passonneau.
  2023.
\newblock Survey on sociodemographic bias in natural language processing.
\newblock \emph{arXiv preprint arXiv:2306.08158}.

\bibitem[{Han et~al.(2018)Han, Zhang, Zhang, Yang, and Zou}]{han2018improving}
Hongyu Han, Yongshi Zhang, Jianpei Zhang, Jing Yang, and Xiaomei Zou. 2018.
\newblock Improving the performance of lexicon-based review sentiment analysis
  method by reducing additional introduced sentiment bias.
\newblock \emph{PLoS One}, 13(8):e0202523.

\bibitem[{Hannak et~al.(2012)Hannak, Anderson, Barrett, Lehmann, Mislove, and
  Riedewald}]{hannak2012tweetin}
Aniko Hannak, Eric Anderson, Lisa~Feldman Barrett, Sune Lehmann, Alan Mislove,
  and Mirek Riedewald. 2012.
\newblock Tweetin'in the rain: Exploring societal-scale effects of weather on
  mood.
\newblock In \emph{Proceedings of the International AAAI Conference on Web and
  Social Media}, volume~6, pages 479--482.

\bibitem[{Hassan et~al.(2017)Hassan, Hussain, Hussain, Sadiq, and
  Lee}]{hassan2017sentiment}
Anees~Ul Hassan, Jamil Hussain, Musarrat Hussain, Muhammad Sadiq, and Sungyoung
  Lee. 2017.
\newblock Sentiment analysis of social networking sites (sns) data using
  machine learning approach for the measurement of depression.
\newblock In \emph{2017 international conference on information and
  communication technology convergence (ICTC)}, pages 138--140. IEEE.

\bibitem[{He and McAuley(2016)}]{he2016ups}
Ruining He and Julian McAuley. 2016.
\newblock Ups and downs: Modeling the visual evolution of fashion trends with
  one-class collaborative filtering.
\newblock In \emph{proceedings of the 25th international conference on world
  wide web}, pages 507--517.

\bibitem[{Head et~al.(2015)Head, Holman, Lanfear, Kahn, and
  Jennions}]{head2015extent}
Megan~L Head, Luke Holman, Rob Lanfear, Andrew~T Kahn, and Michael~D Jennions.
  2015.
\newblock The extent and consequences of p-hacking in science.
\newblock \emph{PLoS biology}, 13(3):e1002106.

\bibitem[{Hoffmann(2018)}]{hoffmann2018too}
Thomas Hoffmann. 2018.
\newblock “too many americans are trapped in fear, violence and poverty”: a
  psychology-informed sentiment analysis of campaign speeches from the 2016 us
  presidential election.
\newblock \emph{Linguistics Vanguard}, 4(1).

\bibitem[{Hu and Liu(2004)}]{hu2004mining}
Minqing Hu and Bing Liu. 2004.
\newblock Mining and summarizing customer reviews.
\newblock In \emph{Proceedings of the tenth ACM SIGKDD international conference
  on Knowledge discovery and data mining}, pages 168--177.

\bibitem[{Huang et~al.(2020)Huang, Zhang, Jiang, Stanforth, Welbl, Rae, Maini,
  Yogatama, and Kohli}]{huang_reducing_2020}
Po-Sen Huang, Huan Zhang, Ray Jiang, Robert Stanforth, Johannes Welbl, Jack
  Rae, Vishal Maini, Dani Yogatama, and Pushmeet Kohli. 2020.
\newblock \href {https://doi.org/10.18653/v1/2020.findings-emnlp.7} {Reducing
  {Sentiment} {Bias} in {Language} {Models} via {Counterfactual} {Evaluation}}.
\newblock In \emph{Findings of the {Association} for {Computational}
  {Linguistics}: {EMNLP} 2020}, pages 65--83. Association for Computational
  Linguistics.

\bibitem[{Hube et~al.(2020)Hube, Idahl, and Fetahu}]{hube_debiasing_2020}
Christoph Hube, Maximilian Idahl, and Besnik Fetahu. 2020.
\newblock \href {https://doi.org/10.1145/3336191.3371779} {Debiasing word
  embeddings from sentiment associations in names}.
\newblock In \emph{Proceedings of the 13th International Conference on Web
  Search and Data Mining}, WSDM '20, page 259–267, New York, NY, USA.
  Association for Computing Machinery.

\bibitem[{Hutchinson et~al.(2020)Hutchinson, Prabhakaran, Denton, Webster,
  Zhong, and Denuyl}]{hutchinson-etal-2020-social}
Ben Hutchinson, Vinodkumar Prabhakaran, Emily Denton, Kellie Webster, Yu~Zhong,
  and Stephen Denuyl. 2020.
\newblock \href {https://doi.org/10.18653/v1/2020.acl-main.487} {Social biases
  in {NLP} models as barriers for persons with disabilities}.
\newblock In \emph{Proceedings of the 58th Annual Meeting of the Association
  for Computational Linguistics}, pages 5491--5501, Online. Association for
  Computational Linguistics.

\bibitem[{Hutto and Gilbert(2014)}]{hutto2014vader}
Clayton Hutto and Eric Gilbert. 2014.
\newblock Vader: A parsimonious rule-based model for sentiment analysis of
  social media text.
\newblock In \emph{Proceedings of the international AAAI conference on web and
  social media}, volume~8, pages 216--225.

\bibitem[{Ikoro et~al.(2018)Ikoro, Sharmina, Malik, and
  Batista-Navarro}]{ikoro2018analyzing}
Victoria Ikoro, Maria Sharmina, Khaleel Malik, and Riza Batista-Navarro. 2018.
\newblock Analyzing sentiments expressed on twitter by uk energy company
  consumers.
\newblock In \emph{2018 Fifth international conference on social networks
  analysis, management and security (SNAMS)}, pages 95--98. IEEE.

\bibitem[{Ionescu and Butnaru(2019)}]{ionescu2019vector}
Radu~Tudor Ionescu and Andrei~M Butnaru. 2019.
\newblock Vector of locally-aggregated word embeddings (vlawe): A novel
  document-level representation.
\newblock \emph{arXiv preprint arXiv:1902.08850}.

\bibitem[{Isah et~al.(2014)Isah, Trundle, and Neagu}]{isah2014social}
Haruna Isah, Paul Trundle, and Daniel Neagu. 2014.
\newblock Social media analysis for product safety using text mining and
  sentiment analysis.
\newblock In \emph{2014 14th UK workshop on computational intelligence (UKCI)},
  pages 1--7. IEEE.

\bibitem[{Izard(2010)}]{izard2010many}
Carroll~E Izard. 2010.
\newblock The many meanings/aspects of emotion: Definitions, functions,
  activation, and regulation.
\newblock \emph{Emotion Review}, 2(4):363--370.

\bibitem[{Izzo and Maloy(2017)}]{izzo201786}
JA~Izzo and K~Maloy. 2017.
\newblock 86 sentiment analysis demonstrates variability in medical student
  grading.
\newblock \emph{Annals of Emergency Medicine}, 70(4):S35--S36.

\bibitem[{Jabbar et~al.(2019)Jabbar, Urooj, JunSheng, and
  Azeem}]{jabbar2019real}
Jahanzeb Jabbar, Iqra Urooj, Wu~JunSheng, and Naqash Azeem. 2019.
\newblock Real-time sentiment analysis on e-commerce application.
\newblock In \emph{2019 IEEE 16th international conference on networking,
  sensing and control (ICNSC)}, pages 391--396. IEEE.

\bibitem[{Ji et~al.(2013)Ji, Chun, and Geller}]{ji2013monitoring}
Xiang Ji, Soon~Ae Chun, and James Geller. 2013.
\newblock Monitoring public health concerns using twitter sentiment
  classifications.
\newblock In \emph{2013 IEEE International Conference on Healthcare
  Informatics}, pages 335--344. IEEE.

\bibitem[{Jing and Murugesan(2019)}]{jing2019theoretical}
Tee~Wee Jing and Raja~Kumar Murugesan. 2019.
\newblock A theoretical framework to build trust and prevent fake news in
  social media using blockchain.
\newblock In \emph{Recent Trends in Data Science and Soft Computing:
  Proceedings of the 3rd International Conference of Reliable Information and
  Communication Technology (IRICT 2018)}, pages 955--962. Springer.

\bibitem[{Jo et~al.(2017)Jo, Kim, and Ryu}]{jo2017we}
Hwiyeol Jo, Soo-Min Kim, and Jeong Ryu. 2017.
\newblock What we really want to find by sentiment analysis: the relationship
  between computational models and psychological state.
\newblock \emph{arXiv preprint arXiv:1704.03407}.

\bibitem[{Joyce and Deng(2017)}]{joyce2017sentiment}
Brandon Joyce and Jing Deng. 2017.
\newblock Sentiment analysis of tweets for the 2016 us presidential election.
\newblock In \emph{2017 ieee mit undergraduate research technology conference
  (urtc)}, pages 1--4. IEEE.

\bibitem[{Kenyon-Dean et~al.(2018)Kenyon-Dean, Ahmed, Fujimoto,
  Georges-Filteau, Glasz, Kaur, Lalande, Bhanderi, Belfer, Kanagasabai
  et~al.}]{kenyon2018sentiment}
Kian Kenyon-Dean, Eisha Ahmed, Scott Fujimoto, Jeremy Georges-Filteau,
  Christopher Glasz, Barleen Kaur, Auguste Lalande, Shruti Bhanderi, Robert
  Belfer, Nirmal Kanagasabai, et~al. 2018.
\newblock Sentiment analysis: It’s complicated!
\newblock In \emph{Proceedings of the 2018 Conference of the North American
  Chapter of the Association for Computational Linguistics: Human Language
  Technologies, Volume 1 (Long Papers)}, pages 1886--1895.

\bibitem[{Kiritchenko and Mohammad(2018)}]{kiritchenko-mohammad-2018-examining}
Svetlana Kiritchenko and Saif Mohammad. 2018.
\newblock \href {https://doi.org/10.18653/v1/S18-2005} {Examining gender and
  race bias in two hundred sentiment analysis systems}.
\newblock In \emph{Proceedings of the Seventh Joint Conference on Lexical and
  Computational Semantics}, pages 43--53, New Orleans, Louisiana. Association
  for Computational Linguistics.

\bibitem[{Korkontzelos et~al.(2016)Korkontzelos, Nikfarjam, Shardlow, Sarker,
  Ananiadou, and Gonzalez}]{korkontzelos2016analysis}
Ioannis Korkontzelos, Azadeh Nikfarjam, Matthew Shardlow, Abeed Sarker, Sophia
  Ananiadou, and Graciela~H Gonzalez. 2016.
\newblock Analysis of the effect of sentiment analysis on extracting adverse
  drug reactions from tweets and forum posts.
\newblock \emph{Journal of biomedical informatics}, 62:148--158.

\bibitem[{Kothari et~al.(2020)Kothari, Shah, Khara, and
  Prajapati}]{Kothari2020ANA}
Komal Kothari, Aagam Shah, Satvik Khara, and Himanshu Prajapati. 2020.
\newblock A novel approach in user reviews analysis using text summarization
  and sentiment analysis: Survey.

\bibitem[{Kraaijeveld and De~Smedt(2020)}]{kraaijeveld2020predictive}
Olivier Kraaijeveld and Johannes De~Smedt. 2020.
\newblock The predictive power of public twitter sentiment for forecasting
  cryptocurrency prices.
\newblock \emph{Journal of International Financial Markets, Institutions and
  Money}, 65:101188.

\bibitem[{Kwon et~al.(2006)Kwon, Shulman, and Hovy}]{kwon2006multidimensional}
Namhee Kwon, Stuart~W Shulman, and Eduard Hovy. 2006.
\newblock Multidimensional text analysis for erulemaking.
\newblock In \emph{Proceedings of the 2006 international conference on Digital
  government research}, pages 157--166.

\bibitem[{Laskari and Sanampudi(2016)}]{laskari2016aspect}
Naveen~Kumar Laskari and Suresh~Kumar Sanampudi. 2016.
\newblock Aspect based sentiment analysis survey.
\newblock \emph{IOSR Journal of Computer Engineering (IOSR-JCE)}, 18(2):24--28.

\bibitem[{Lewicki et~al.(2023)Lewicki, Lee, Cobbe, and Singh}]{lewicki2023out}
Kornel Lewicki, Michelle Seng~Ah Lee, Jennifer Cobbe, and Jatinder Singh. 2023.
\newblock Out of context: Investigating the bias and fairness concerns of"
  artificial intelligence as a service".
\newblock \emph{arXiv preprint arXiv:2302.01448}.

\bibitem[{Li et~al.(2023)Li, Bruce, Li, and Gao}]{li2023restaurant}
Hengyun Li, XB~Bruce, Gang Li, and Huicai Gao. 2023.
\newblock Restaurant survival prediction using customer-generated content: An
  aspect-based sentiment analysis of online reviews.
\newblock \emph{Tourism Management}, 96:104707.

\bibitem[{Li et~al.(2016)Li, Cui, Shen, and Ma}]{li2016intelligent}
Hui Li, Jiangtao Cui, Bingqing Shen, and Jianfeng Ma. 2016.
\newblock An intelligent movie recommendation system through group-level
  sentiment analysis in microblogs.
\newblock \emph{Neurocomputing}, 210:164--173.

\bibitem[{Li et~al.(2018)Li, Ebrahimi~Kahou, Schulz, Michalski, Charlin, and
  Pal}]{li2018towards}
Raymond Li, Samira Ebrahimi~Kahou, Hannes Schulz, Vincent Michalski, Laurent
  Charlin, and Chris Pal. 2018.
\newblock Towards deep conversational recommendations.
\newblock \emph{Advances in neural information processing systems}, 31.

\bibitem[{Lin et~al.(2021)Lin, Liu, Xv, and Li}]{lin2021mitigating}
Chen Lin, Xinyi Liu, Guipeng Xv, and Hui Li. 2021.
\newblock Mitigating sentiment bias for recommender systems.
\newblock In \emph{Proceedings of the 44th International ACM SIGIR Conference
  on Research and Development in Information Retrieval}, pages 31--40.

\bibitem[{Liu(2012)}]{liu2012sentiment}
Bing Liu. 2012.
\newblock Sentiment analysis and opinion mining.
\newblock \emph{Synthesis lectures on human language technologies},
  5(1):1--167.

\bibitem[{Lou et~al.(2020)Lou, Zhang, Li, Qian, and Ji}]{lou2020emoji}
Yinxia Lou, Yue Zhang, Fei Li, Tao Qian, and Donghong Ji. 2020.
\newblock Emoji-based sentiment analysis using attention networks.
\newblock \emph{ACM Transactions on asian and low-resource language information
  processing (TALLIP)}, 19(5):1--13.

\bibitem[{Loughran and McDonald(2011)}]{loughran2011liability}
Tim Loughran and Bill McDonald. 2011.
\newblock When is a liability not a liability? textual analysis, dictionaries,
  and 10-ks.
\newblock \emph{The Journal of finance}, 66(1):35--65.

\bibitem[{Lyu et~al.(2020)Lyu, Foster, and Graham}]{lyu2020improving}
Chenyang Lyu, Jennifer Foster, and Yvette Graham. 2020.
\newblock Improving document-level sentiment analysis with user and product
  context.
\newblock \emph{arXiv preprint arXiv:2011.09210}.

\bibitem[{Ma et~al.(2017)Ma, Li, Zhang, Wang, and Sun}]{ma2017cascading}
Dehong Ma, Sujian Li, Xiaodong Zhang, Houfeng Wang, and Xu~Sun. 2017.
\newblock Cascading multiway attentions for document-level sentiment
  classification.
\newblock In \emph{Proceedings of the Eighth International Joint Conference on
  Natural Language Processing (Volume 1: Long Papers)}, pages 634--643.

\bibitem[{Ma et~al.(2018)Ma, Peng, and Cambria}]{ma2018targeted}
Yukun Ma, Haiyun Peng, and Erik Cambria. 2018.
\newblock Targeted aspect-based sentiment analysis via embedding commonsense
  knowledge into an attentive lstm.
\newblock In \emph{Proceedings of the AAAI conference on artificial
  intelligence}, volume~32.

\bibitem[{Maas et~al.(2011)Maas, Daly, Pham, Huang, Ng, and
  Potts}]{maas2011learning}
Andrew Maas, Raymond~E Daly, Peter~T Pham, Dan Huang, Andrew~Y Ng, and
  Christopher Potts. 2011.
\newblock Learning word vectors for sentiment analysis.
\newblock In \emph{Proceedings of the 49th annual meeting of the association
  for computational linguistics: Human language technologies}, pages 142--150.

\bibitem[{Mahtab et~al.(2018)Mahtab, Islam, and Rahaman}]{mahtab2018sentiment}
Shamsul~Arafin Mahtab, Nazmul Islam, and Md~Mahfuzur Rahaman. 2018.
\newblock Sentiment analysis on bangladesh cricket with support vector machine.
\newblock In \emph{2018 international conference on Bangla speech and language
  processing (ICBSLP)}, pages 1--4. IEEE.

\bibitem[{Majumder et~al.(2019)Majumder, Li, Ni, and
  McAuley}]{majumder2019generating}
Bodhisattwa~Prasad Majumder, Shuyang Li, Jianmo Ni, and Julian McAuley. 2019.
\newblock Generating personalized recipes from historical user preferences.
\newblock \emph{arXiv preprint arXiv:1909.00105}.

\bibitem[{Mansour(2018)}]{mansour2018social}
Samah Mansour. 2018.
\newblock Social media analysis of user’s responses to terrorism using
  sentiment analysis and text mining.
\newblock \emph{Procedia Computer Science}, 140:95--103.

\bibitem[{Medhat et~al.(2014)Medhat, Hassan, and Korashy}]{medhat2014sentiment}
Walaa Medhat, Ahmed Hassan, and Hoda Korashy. 2014.
\newblock Sentiment analysis algorithms and applications: A survey.
\newblock \emph{Ain Shams engineering journal}, 5(4):1093--1113.

\bibitem[{Mei et~al.(2023)Mei, Fereidooni, and Caliskan}]{mei2023bias}
Katelyn Mei, Sonia Fereidooni, and Aylin Caliskan. 2023.
\newblock Bias against 93 stigmatized groups in masked language models and
  downstream sentiment classification tasks.
\newblock In \emph{Proceedings of the 2023 ACM Conference on Fairness,
  Accountability, and Transparency}, pages 1699--1710.

\bibitem[{Mohammad(2012)}]{mohammad2012emotional}
Saif Mohammad. 2012.
\newblock \# emotional tweets.
\newblock In \emph{* SEM 2012: The First Joint Conference on Lexical and
  Computational Semantics--Volume 1: Proceedings of the main conference and the
  shared task, and Volume 2: Proceedings of the Sixth International Workshop on
  Semantic Evaluation (SemEval 2012)}, pages 246--255.

\bibitem[{Mohammad(2022)}]{mohammad2022ethics}
Saif~M Mohammad. 2022.
\newblock Ethics sheet for automatic emotion recognition and sentiment
  analysis.
\newblock \emph{Computational Linguistics}, 48(2):239--278.

\bibitem[{Mohammad et~al.(2013)Mohammad, Kiritchenko, and
  Zhu}]{mohammad2013nrc}
Saif~M Mohammad, Svetlana Kiritchenko, and Xiaodan Zhu. 2013.
\newblock Nrc-canada: Building the state-of-the-art in sentiment analysis of
  tweets.
\newblock \emph{arXiv preprint arXiv:1308.6242}.

\bibitem[{Mohammad and Turney(2013)}]{mohammad2013crowdsourcing}
Saif~M Mohammad and Peter~D Turney. 2013.
\newblock Crowdsourcing a word--emotion association lexicon.
\newblock \emph{Computational intelligence}, 29(3):436--465.

\bibitem[{Moreno-Ortiz and P{\'e}rez-Hern{\'a}ndez(2018)}]{moreno2018lingmotif}
Antonio Moreno-Ortiz and Chantal P{\'e}rez-Hern{\'a}ndez. 2018.
\newblock Lingmotif-lex: a wide-coverage, state-of-the-art lexicon for
  sentiment analysis.
\newblock In \emph{Proceedings of the Eleventh International Conference on
  Language Resources and Evaluation (LREC 2018)}.

\bibitem[{Mowlaei et~al.(2020)Mowlaei, Abadeh, and
  Keshavarz}]{mowlaei2020aspect}
Mohammad~Erfan Mowlaei, Mohammad~Saniee Abadeh, and Hamidreza Keshavarz. 2020.
\newblock Aspect-based sentiment analysis using adaptive aspect-based lexicons.
\newblock \emph{Expert Systems with Applications}, 148:113234.

\bibitem[{Nakov et~al.(2013)Nakov, Rosenthal, Kozareva, Stoyanov, Ritter, and
  Wilson}]{nakov-etal-2013-semeval}
Preslav Nakov, Sara Rosenthal, Zornitsa Kozareva, Veselin Stoyanov, Alan
  Ritter, and Theresa Wilson. 2013.
\newblock \href {https://aclanthology.org/S13-2052} {{S}em{E}val-2013 task 2:
  Sentiment analysis in {T}witter}.
\newblock In \emph{Second Joint Conference on Lexical and Computational
  Semantics (*{SEM}), Volume 2: Proceedings of the Seventh International
  Workshop on Semantic Evaluation ({S}em{E}val 2013)}, pages 312--320, Atlanta,
  Georgia, USA. Association for Computational Linguistics.

\bibitem[{Narayanan~Venkit et~al.(2023)Narayanan~Venkit, Gautam, Panchanadikar,
  Huang, and Wilson}]{narayanan2023unmasking}
Pranav Narayanan~Venkit, Sanjana Gautam, Ruchi Panchanadikar, Ting-Hao Huang,
  and Shomir Wilson. 2023.
\newblock Unmasking nationality bias: A study of human perception of
  nationalities in ai-generated articles.
\newblock In \emph{Proceedings of the 2023 AAAI/ACM Conference on AI, Ethics,
  and Society}, pages 554--565.

\bibitem[{Nielsen(2011)}]{nielsen2011new}
Finn~{\AA}rup Nielsen. 2011.
\newblock A new anew: Evaluation of a word list for sentiment analysis in
  microblogs.
\newblock \emph{arXiv preprint arXiv:1103.2903}.

\bibitem[{O'neil(2017)}]{o2017weapons}
Cathy O'neil. 2017.
\newblock \emph{Weapons of math destruction: How big data increases inequality
  and threatens democracy}.
\newblock Crown.

\bibitem[{Ortigosa et~al.(2014)Ortigosa, Mart{\'\i}n, and
  Carro}]{ortigosa2014sentiment}
Alvaro Ortigosa, Jos{\'e}~M Mart{\'\i}n, and Rosa~M Carro. 2014.
\newblock Sentiment analysis in facebook and its application to e-learning.
\newblock \emph{Computers in human behavior}, 31:527--541.

\bibitem[{Pang and Lee(2004)}]{pang2004sentimental}
Bo~Pang and Lillian Lee. 2004.
\newblock A sentimental education: Sentiment analysis using subjectivity
  summarization based on minimum cuts.
\newblock \emph{arXiv preprint cs/0409058}.

\bibitem[{Pang and Lee(2005)}]{pang2005seeing}
Bo~Pang and Lillian Lee. 2005.
\newblock Seeing stars: Exploiting class relationships for sentiment
  categorization with respect to rating scales.
\newblock \emph{arXiv preprint cs/0506075}.

\bibitem[{Pang et~al.(2002)Pang, Lee, and Vaithyanathan}]{pang2002thumbs}
Bo~Pang, Lillian Lee, and Shivakumar Vaithyanathan. 2002.
\newblock Thumbs up? sentiment classification using machine learning
  techniques.
\newblock \emph{arXiv preprint cs/0205070}.

\bibitem[{Pappas et~al.(2013)Pappas, Katsimpras, and
  Stamatatos}]{pappas2013distinguishing}
Nikolaos Pappas, Georgios Katsimpras, and Efstathios Stamatatos. 2013.
\newblock Distinguishing the popularity between topics: a system for up-to-date
  opinion retrieval and mining in the web.
\newblock In \emph{Computational Linguistics and Intelligent Text Processing:
  14th International Conference, CICLing 2013, Samos, Greece, March 24-30,
  2013, Proceedings, Part II 14}, pages 197--209. Springer.

\bibitem[{Plutchik(2001)}]{plutchik2001nature}
Robert Plutchik. 2001.
\newblock The nature of emotions: Human emotions have deep evolutionary roots,
  a fact that may explain their complexity and provide tools for clinical
  practice.
\newblock \emph{American scientist}, 89(4):344--350.

\bibitem[{Prabhakaran et~al.(2019)Prabhakaran, Hutchinson, and
  Mitchell}]{prabhakaran-etal-2019-perturbation}
Vinodkumar Prabhakaran, Ben Hutchinson, and Margaret Mitchell. 2019.
\newblock \href {https://doi.org/10.18653/v1/D19-1578} {Perturbation
  sensitivity analysis to detect unintended model biases}.
\newblock In \emph{Proceedings of the 2019 Conference on Empirical Methods in
  Natural Language Processing and the 9th International Joint Conference on
  Natural Language Processing (EMNLP-IJCNLP)}, pages 5740--5745, Hong Kong,
  China. Association for Computational Linguistics.

\bibitem[{Price et~al.(2012)Price, Doran, Peterson, and
  Bliss}]{price2012earnings}
S~McKay Price, James~S Doran, David~R Peterson, and Barbara~A Bliss. 2012.
\newblock Earnings conference calls and stock returns: The incremental
  informativeness of textual tone.
\newblock \emph{Journal of Banking \& Finance}, 36(4):992--1011.

\bibitem[{Prun and Raymond(2021)}]{Prun2021ACE}
Daniel Prun and Camille Raymond. 2021.
\newblock A controlled experiment on using cognitive work analysis for system
  engineering definition process.
\newblock \emph{2021 16th International Conference of System of Systems
  Engineering (SoSE)}, pages 1--6.

\bibitem[{Raffel et~al.(2020)Raffel, Shazeer, Roberts, Lee, Narang, Matena,
  Zhou, Li, and Liu}]{raffel2020exploring}
Colin Raffel, Noam Shazeer, Adam Roberts, Katherine Lee, Sharan Narang, Michael
  Matena, Yanqi Zhou, Wei Li, and Peter~J Liu. 2020.
\newblock Exploring the limits of transfer learning with a unified text-to-text
  transformer.
\newblock \emph{The Journal of Machine Learning Research}, 21(1):5485--5551.

\bibitem[{Rajput(2020)}]{rajput2020natural}
Adil Rajput. 2020.
\newblock Natural language processing, sentiment analysis, and clinical
  analytics.
\newblock In \emph{Innovation in health informatics}, pages 79--97. Elsevier.

\bibitem[{Ram{\'\i}rez-Tinoco et~al.(2019)Ram{\'\i}rez-Tinoco,
  Alor-Hern{\'a}ndez, S{\'a}nchez-Cervantes, Salas-Z{\'a}rate, and
  Valencia-Garc{\'\i}a}]{ramirez2019use}
Francisco~Javier Ram{\'\i}rez-Tinoco, Giner Alor-Hern{\'a}ndez, Jos{\'e}~Luis
  S{\'a}nchez-Cervantes, Mar{\'\i}a del~Pilar Salas-Z{\'a}rate, and Rafael
  Valencia-Garc{\'\i}a. 2019.
\newblock Use of sentiment analysis techniques in healthcare domain.
\newblock \emph{Current Trends in Semantic Web Technologies: Theory and
  Practice}, pages 189--212.

\bibitem[{Rani et~al.(2022)Rani, Bashir, Alhudhaif, Koundal, Gunduz
  et~al.}]{rani2022efficient}
Shalli Rani, Ali~Kashif Bashir, Adi Alhudhaif, Deepika Koundal, Emine~Selda
  Gunduz, et~al. 2022.
\newblock An efficient cnn-lstm model for sentiment detection in\#
  blacklivesmatter.
\newblock \emph{Expert Systems with Applications}, 193:116256.

\bibitem[{Read(2005)}]{read2005using}
Jonathon Read. 2005.
\newblock Using emoticons to reduce dependency in machine learning techniques
  for sentiment classification.
\newblock In \emph{Proceedings of the ACL student research workshop}, pages
  43--48.

\bibitem[{Ribeiro et~al.(2016)Ribeiro, Ara{\'u}jo, Gon{\c{c}}alves,
  Andr{\'e}~Gon{\c{c}}alves, and Benevenuto}]{ribeiro2016sentibench}
Filipe~N Ribeiro, Matheus Ara{\'u}jo, Pollyanna Gon{\c{c}}alves, Marcos
  Andr{\'e}~Gon{\c{c}}alves, and Fabr{\'\i}cio Benevenuto. 2016.
\newblock Sentibench-a benchmark comparison of state-of-the-practice sentiment
  analysis methods.
\newblock \emph{EPJ Data Science}, 5:1--29.

\bibitem[{Rietzler et~al.(2019)Rietzler, Stabinger, Opitz, and
  Engl}]{rietzler2019adapt}
Alexander Rietzler, Sebastian Stabinger, Paul Opitz, and Stefan Engl. 2019.
\newblock Adapt or get left behind: Domain adaptation through bert language
  model finetuning for aspect-target sentiment classification.
\newblock \emph{arXiv preprint arXiv:1908.11860}.

\bibitem[{Rodrigues et~al.(2016)Rodrigues, das Dores, Camilo-Junior, and
  Rosa}]{rodrigues2016sentihealth}
Ramon~Gouveia Rodrigues, Rafael~Marques das Dores, Celso~G Camilo-Junior, and
  Thierson~Couto Rosa. 2016.
\newblock Sentihealth-cancer: a sentiment analysis tool to help detecting mood
  of patients in online social networks.
\newblock \emph{International journal of medical informatics}, 85(1):80--95.

\bibitem[{Rognone et~al.(2020)Rognone, Hyde, and Zhang}]{rognone2020news}
Lavinia Rognone, Stuart Hyde, and S~Sarah Zhang. 2020.
\newblock News sentiment in the cryptocurrency market: An empirical comparison
  with forex.
\newblock \emph{International Review of Financial Analysis}, 69:101462.

\bibitem[{Rosenthal et~al.(2019)Rosenthal, Farra, and
  Nakov}]{rosenthal2019semeval}
Sara Rosenthal, Noura Farra, and Preslav Nakov. 2019.
\newblock Semeval-2017 task 4: Sentiment analysis in twitter.
\newblock \emph{arXiv preprint arXiv:1912.00741}.

\bibitem[{Rozado(2020)}]{rozado2020wide}
David Rozado. 2020.
\newblock Wide range screening of algorithmic bias in word embedding models
  using large sentiment lexicons reveals underreported bias types.
\newblock \emph{PloS one}, 15(4):e0231189.

\bibitem[{Rudin(2019)}]{rudin2019stop}
Cynthia Rudin. 2019.
\newblock Stop explaining black box machine learning models for high stakes
  decisions and use interpretable models instead.
\newblock \emph{Nature machine intelligence}, 1(5):206--215.

\bibitem[{Sabra et~al.(2018)Sabra, Malik, and Alobaidi}]{sabra2018prediction}
Susan Sabra, Khalid~Mahmood Malik, and Mazen Alobaidi. 2018.
\newblock Prediction of venous thromboembolism using semantic and sentiment
  analyses of clinical narratives.
\newblock \emph{Computers in biology and medicine}, 94:1--10.

\bibitem[{Saeidi et~al.(2016)Saeidi, Bouchard, Liakata, and
  Riedel}]{saeidi2016sentihood}
Marzieh Saeidi, Guillaume Bouchard, Maria Liakata, and Sebastian Riedel. 2016.
\newblock Sentihood: Targeted aspect based sentiment analysis dataset for urban
  neighbourhoods.
\newblock \emph{arXiv preprint arXiv:1610.03771}.

\bibitem[{Salas-Z{\'a}rate et~al.(2017)Salas-Z{\'a}rate, Medina-Moreira,
  Lagos-Ortiz, Luna-Aveiga, Rodriguez-Garcia, and
  Valencia-Garcia}]{salas2017sentiment}
Mar{\'\i}a del~Pilar Salas-Z{\'a}rate, Jose Medina-Moreira, Katty Lagos-Ortiz,
  Harry Luna-Aveiga, Miguel~Angel Rodriguez-Garcia, and Rafael Valencia-Garcia.
  2017.
\newblock Sentiment analysis on tweets about diabetes: an aspect-level
  approach.
\newblock \emph{Computational and mathematical methods in medicine}, 2017.

\bibitem[{S{\'a}nchez-Rada and Iglesias(2019)}]{sanchez2019social}
J~Fernando S{\'a}nchez-Rada and Carlos~A Iglesias. 2019.
\newblock Social context in sentiment analysis: Formal definition, overview of
  current trends and framework for comparison.
\newblock \emph{Information Fusion}, 52:344--356.

\bibitem[{Saragih and Girsang(2017)}]{saragih2017sentiment}
Melva~Hermayanty Saragih and Abba~Suganda Girsang. 2017.
\newblock Sentiment analysis of customer engagement on social media in
  transport online.
\newblock In \emph{2017 International Conference on Sustainable Information
  Engineering and Technology (SIET)}, pages 24--29. IEEE.

\bibitem[{Schumaker et~al.(2012)Schumaker, Zhang, Huang, and
  Chen}]{schumaker2012evaluating}
Robert~P Schumaker, Yulei Zhang, Chun-Neng Huang, and Hsinchun Chen. 2012.
\newblock Evaluating sentiment in financial news articles.
\newblock \emph{Decision Support Systems}, 53(3):458--464.

\bibitem[{Schwartz et~al.(2021)Schwartz, Down, Jonas, and
  Tabassi}]{schwartz2021proposal}
Reva Schwartz, Leann Down, Adam Jonas, and Elham Tabassi. 2021.
\newblock A proposal for identifying and managing bias in artificial
  intelligence.
\newblock \emph{Draft NIST Special Publication}, 1270.

\bibitem[{Sebastiani and Esuli(2006)}]{sebastiani2006sentiwordnet}
Fabrizio Sebastiani and Andrea Esuli. 2006.
\newblock Sentiwordnet: A publicly available lexical resource for opinion
  mining.
\newblock In \emph{Proceedings of the 5th international conference on language
  resources and evaluation}, pages 417--422. European Language Resources
  Association (ELRA) Genoa, Italy.

\bibitem[{Shahare(2017)}]{shahare2017sentiment}
Firoj~Fattulal Shahare. 2017.
\newblock Sentiment analysis for the news data based on the social media.
\newblock In \emph{2017 International Conference on Intelligent Computing and
  Control Systems (ICICCS)}, pages 1365--1370. IEEE.

\bibitem[{Shayaa et~al.(2017)Shayaa, Wai, Chung, Sulaiman, Jaafar, and
  Zakaria}]{shayaa2017social}
Shahid Shayaa, Phoong~Seuk Wai, Yeong~Wai Chung, Ainin Sulaiman, Noor~Ismawati
  Jaafar, and Shamshul~Bahri Zakaria. 2017.
\newblock Social media sentiment analysis on employment in malaysia.
\newblock In \emph{the Proceedings of 8th Global Business and Finance Research
  Conference, Taipei, Taiwan}.

\bibitem[{Shen et~al.(2018)Shen, Fratamico, Rahwan, and Rush}]{shen2018darling}
Judy~Hanwen Shen, Lauren Fratamico, Iyad Rahwan, and Alexander~M Rush. 2018.
\newblock Darling or babygirl? investigating stylistic bias in sentiment
  analysis.
\newblock \emph{Proc. of FATML}.

\bibitem[{Socher et~al.(2013)Socher, Perelygin, Wu, Chuang, Manning, Ng, and
  Potts}]{socher2013recursive}
Richard Socher, Alex Perelygin, Jean Wu, Jason Chuang, Christopher~D Manning,
  Andrew~Y Ng, and Christopher Potts. 2013.
\newblock Recursive deep models for semantic compositionality over a sentiment
  treebank.
\newblock In \emph{Proceedings of the 2013 conference on empirical methods in
  natural language processing}, pages 1631--1642.

\bibitem[{Soni and Rambola(2022)}]{soni2022survey}
Piyush~Kumar Soni and Radhakrishna Rambola. 2022.
\newblock A survey on implicit aspect detection for sentiment analysis:
  terminology, issues, and scope.
\newblock \emph{IEEE Access}, 10:63932--63957.

\bibitem[{Stark and Hoey(2021)}]{stark2021ethics}
Luke Stark and Jesse Hoey. 2021.
\newblock The ethics of emotion in artificial intelligence systems.
\newblock In \emph{Proceedings of the 2021 ACM Conference on Fairness,
  Accountability, and Transparency}, pages 782--793.

\bibitem[{Sun et~al.(2019)Sun, Huang, and Qiu}]{sun2019utilizing}
Chi Sun, Luyao Huang, and Xipeng Qiu. 2019.
\newblock Utilizing bert for aspect-based sentiment analysis via constructing
  auxiliary sentence.
\newblock \emph{arXiv preprint arXiv:1903.09588}.

\bibitem[{Sweeney and Najafian(2020)}]{sweeney2020reducing}
Chris Sweeney and Maryam Najafian. 2020.
\newblock Reducing sentiment polarity for demographic attributes in word
  embeddings using adversarial learning.
\newblock In \emph{Proceedings of the 2020 Conference on Fairness,
  Accountability, and Transparency}, pages 359--368.

\bibitem[{Syaifudin and Puspitasari(2017)}]{syaifudin2017twitter}
Y~Watequlis Syaifudin and Dwi Puspitasari. 2017.
\newblock Twitter data mining for sentiment analysis on peoples feedback
  against government public policy.
\newblock \emph{Matter Int. J. Sci. Technol}, 3(1):110--122.

\bibitem[{Taboada(2016)}]{taboada2016sentiment}
Maite Taboada. 2016.
\newblock Sentiment analysis: An overview from linguistics.
\newblock \emph{Annual Review of Linguistics}, 2:325--347.

\bibitem[{Taboada et~al.(2011)Taboada, Brooke, Tofiloski, Voll, and
  Stede}]{taboada2011lexicon}
Maite Taboada, Julian Brooke, Milan Tofiloski, Kimberly Voll, and Manfred
  Stede. 2011.
\newblock Lexicon-based methods for sentiment analysis.
\newblock \emph{Computational linguistics}, 37(2):267--307.

\bibitem[{Thelwall(2014)}]{thelwall2014heart}
Mike Thelwall. 2014.
\newblock Heart and soul: Sentiment strength detection in the social web with
  sentistrength, 2017.
\newblock \emph{Cyberemotions: Collective emotions in cyberspace}.

\bibitem[{Trist(1981)}]{trist1981evolution}
Eric~L Trist. 1981.
\newblock \emph{The evolution of socio-technical systems}, volume~2.
\newblock Ontario Quality of Working Life Centre Toronto.

\bibitem[{Turney and Littman(2002)}]{turney2002unsupervised}
Peter~D Turney and Michael~L Littman. 2002.
\newblock Unsupervised learning of semantic orientation from a
  hundred-billion-word corpus.
\newblock \emph{arXiv preprint cs/0212012}.

\bibitem[{Ungless et~al.(2023)Ungless, Ross, and Belle}]{ungless2023potential}
Eddie~L Ungless, Bj{\"o}rn Ross, and Vaishak Belle. 2023.
\newblock Potential pitfalls with automatic sentiment analysis: The example of
  queerphobic bias.
\newblock \emph{Social Science Computer Review}, page 08944393231152946.

\bibitem[{Vaidis and Bran(2019)}]{vaidis2019respectable}
David~C Vaidis and Alexandre Bran. 2019.
\newblock Respectable challenges to respectable theory: Cognitive dissonance
  theory requires conceptualization clarification and operational tools.
\newblock \emph{Frontiers in psychology}, 10:1189.

\bibitem[{Vaismoradi et~al.(2013)Vaismoradi, Turunen, and
  Bondas}]{vaismoradi2013content}
Mojtaba Vaismoradi, Hannele Turunen, and Terese Bondas. 2013.
\newblock Content analysis and thematic analysis: Implications for conducting a
  qualitative descriptive study.
\newblock \emph{Nursing \& health sciences}, 15(3):398--405.

\bibitem[{Venkit et~al.(2021)Venkit, Karishma, Hsu, Katiki, Huang, Wilson, and
  Dudas}]{venkit2021asourceful}
Pranav Venkit, Zeba Karishma, Chi-Yang Hsu, Rahul Katiki, Kenneth Huang, Shomir
  Wilson, and Patrick Dudas. 2021.
\newblock Asourceful'twist: Emoji prediction based on sentiment, hashtags and
  application source.
\newblock \emph{arXiv preprint arXiv:2103.07833}.

\bibitem[{Venkit et~al.(2023)Venkit, Srinath, and Wilson}]{venkit2023automated}
Pranav~Narayanan Venkit, Mukund Srinath, and Shomir Wilson. 2023.
\newblock Automated ableism: An exploration of explicit disability biases in
  sentiment and toxicity analysis models.
\newblock \emph{arXiv preprint arXiv:2307.09209}.

\bibitem[{Venkit and Wilson(2021)}]{venkit2021identification}
Pranav~Narayanan Venkit and Shomir Wilson. 2021.
\newblock Identification of bias against people with disabilities in sentiment
  analysis and toxicity detection models.
\newblock \emph{arXiv preprint arXiv:2111.13259}.

\bibitem[{Wang et~al.(2013)Wang, Tsai, Liu, and Chang}]{wang2013financial}
Chuan-Ju Wang, Ming-Feng Tsai, Tse Liu, and Chin-Ting Chang. 2013.
\newblock Financial sentiment analysis for risk prediction.
\newblock In \emph{Proceedings of the Sixth International Joint Conference on
  Natural Language Processing}, pages 802--808.

\bibitem[{Wang et~al.(2012)Wang, Can, Kazemzadeh, Bar, and
  Narayanan}]{wang2012system}
Hao Wang, Do{\u{g}}an Can, Abe Kazemzadeh, Fran{\c{c}}ois Bar, and Shrikanth
  Narayanan. 2012.
\newblock A system for real-time twitter sentiment analysis of 2012 us
  presidential election cycle.
\newblock In \emph{Proceedings of the ACL 2012 system demonstrations}, pages
  115--120.

\bibitem[{Wang et~al.(2022)Wang, Fan, Palacios, Chai, Guetta-Jeanrenaud,
  Obradovich, Zhou, and Zheng}]{wang2022global}
Jianghao Wang, Yichun Fan, Juan Palacios, Yuchen Chai, Nicolas
  Guetta-Jeanrenaud, Nick Obradovich, Chenghu Zhou, and Siqi Zheng. 2022.
\newblock Global evidence of expressed sentiment alterations during the
  covid-19 pandemic.
\newblock \emph{Nature Human Behaviour}, 6(3):349--358.

\bibitem[{Wang et~al.(2016)Wang, Pan, Dahlmeier, and Xiao}]{wang2016recursive}
Wenya Wang, Sinno~Jialin Pan, Daniel Dahlmeier, and Xiaokui Xiao. 2016.
\newblock Recursive neural conditional random fields for aspect-based sentiment
  analysis.
\newblock \emph{arXiv preprint arXiv:1603.06679}.

\bibitem[{Wankhade et~al.(2022)Wankhade, Rao, and
  Kulkarni}]{wankhade2022survey}
Mayur Wankhade, Annavarapu Chandra~Sekhara Rao, and Chaitanya Kulkarni. 2022.
\newblock A survey on sentiment analysis methods, applications, and challenges.
\newblock \emph{Artificial Intelligence Review}, 55(7):5731--5780.

\bibitem[{Warriner et~al.(2013)Warriner, Kuperman, and
  Brysbaert}]{warriner2013norms}
Amy~Beth Warriner, Victor Kuperman, and Marc Brysbaert. 2013.
\newblock Norms of valence, arousal, and dominance for 13,915 english lemmas.
\newblock \emph{Behavior research methods}, 45:1191--1207.

\bibitem[{Wiebe et~al.(2005)Wiebe, Wilson, and Cardie}]{wiebe2005annotating}
Janyce Wiebe, Theresa Wilson, and Claire Cardie. 2005.
\newblock Annotating expressions of opinions and emotions in language.
\newblock \emph{Language resources and evaluation}, 39:165--210.

\bibitem[{Wiebe(1994)}]{wiebe1994tracking}
Janyce~M Wiebe. 1994.
\newblock Tracking point of view in narrative.
\newblock \emph{arXiv preprint cmp-lg/9407019}.

\bibitem[{Wilson et~al.(2005{\natexlab{a}})Wilson, Hoffmann, Somasundaran,
  Kessler, Wiebe, Choi, Cardie, Riloff, and
  Patwardhan}]{wilson2005opinionfinder}
Theresa Wilson, Paul Hoffmann, Swapna Somasundaran, Jason Kessler, Janyce
  Wiebe, Yejin Choi, Claire Cardie, Ellen Riloff, and Siddharth Patwardhan.
  2005{\natexlab{a}}.
\newblock Opinionfinder: A system for subjectivity analysis.
\newblock In \emph{Proceedings of HLT/EMNLP 2005 Interactive Demonstrations},
  pages 34--35.

\bibitem[{Wilson et~al.(2005{\natexlab{b}})Wilson, Wiebe, and
  Hoffmann}]{wilson2005recognizing}
Theresa Wilson, Janyce Wiebe, and Paul Hoffmann. 2005{\natexlab{b}}.
\newblock Recognizing contextual polarity in phrase-level sentiment analysis.
\newblock In \emph{Proceedings of human language technology conference and
  conference on empirical methods in natural language processing}, pages
  347--354.

\bibitem[{Wu et~al.(2015)Wu, Moh, and Khuri}]{wu2015twitter}
Liang Wu, Teng-Sheng Moh, and Natalia Khuri. 2015.
\newblock Twitter opinion mining for adverse drug reactions.
\newblock In \emph{2015 IEEE international conference on big data (Big Data)},
  pages 1570--1574. IEEE.

\bibitem[{Wu and Ong(2021)}]{wu2021context}
Zhengxuan Wu and Desmond~C Ong. 2021.
\newblock Context-guided bert for targeted aspect-based sentiment analysis.
\newblock In \emph{Proceedings of the AAAI conference on artificial
  intelligence}, volume~35, pages 14094--14102.

\bibitem[{Xu et~al.(2019)Xu, Liu, Shu, and Yu}]{xu2019bert}
Hu~Xu, Bing Liu, Lei Shu, and Philip~S Yu. 2019.
\newblock Bert post-training for review reading comprehension and aspect-based
  sentiment analysis.
\newblock \emph{arXiv preprint arXiv:1904.02232}.

\bibitem[{Yang et~al.(2016)Yang, Lee, and Kuo}]{yang2016mining}
Fu-Chen Yang, Anthony~JT Lee, and Sz-Chen Kuo. 2016.
\newblock Mining health social media with sentiment analysis.
\newblock \emph{Journal of medical systems}, 40:1--8.

\bibitem[{Yang et~al.(2019)Yang, Dai, Yang, Carbonell, Salakhutdinov, and
  Le}]{yang2019xlnet}
Zhilin Yang, Zihang Dai, Yiming Yang, Jaime Carbonell, Russ~R Salakhutdinov,
  and Quoc~V Le. 2019.
\newblock Xlnet: Generalized autoregressive pretraining for language
  understanding.
\newblock \emph{Advances in neural information processing systems}, 32.

\bibitem[{Yuliyanti et~al.(2017)Yuliyanti, Djatna, and
  Sukoco}]{yuliyanti2017sentiment}
Siti Yuliyanti, Taufik Djatna, and Heru Sukoco. 2017.
\newblock Sentiment mining of community development program evaluation based on
  social media.
\newblock \emph{TELKOMNIKA (Telecommunication Computing Electronics and
  Control)}, 15(4):1858--1864.

\bibitem[{Zad et~al.(2021)Zad, Heidari, Jones, and Uzuner}]{zad2021survey}
Samira Zad, Maryam Heidari, James~H Jones, and Ozlem Uzuner. 2021.
\newblock A survey on concept-level sentiment analysis techniques of textual
  data.
\newblock In \emph{2021 IEEE World AI IoT Congress (AIIoT)}, pages 0285--0291.
  IEEE.

\bibitem[{Zavattaro et~al.(2015)Zavattaro, French, and
  Mohanty}]{zavattaro2015sentiment}
Staci~M Zavattaro, P~Edward French, and Somya~D Mohanty. 2015.
\newblock A sentiment analysis of us local government tweets: The connection
  between tone and citizen involvement.
\newblock \emph{Government information quarterly}, 32(3):333--341.

\bibitem[{Zhai et~al.(2011)Zhai, Liu, Xu, and Jia}]{zhai2011clustering}
Zhongwu Zhai, Bing Liu, Hua Xu, and Peifa Jia. 2011.
\newblock Clustering product features for opinion mining.
\newblock In \emph{Proceedings of the fourth ACM international conference on
  Web search and data mining}, pages 347--354.

\bibitem[{Zhang and Liu(2011)}]{zhang2011identifying}
Lei Zhang and Bing Liu. 2011.
\newblock Identifying noun product features that imply opinions.
\newblock In \emph{Proceedings of the 49th annual meeting of the Association
  for Computational Linguistics: human language technologies}, pages 575--580.

\bibitem[{Zhang et~al.(2018)Zhang, Wang, and Liu}]{zhang2018deep}
Lei Zhang, Shuai Wang, and Bing Liu. 2018.
\newblock Deep learning for sentiment analysis: A survey.
\newblock \emph{Wiley Interdisciplinary Reviews: Data Mining and Knowledge
  Discovery}, 8(4):e1253.

\bibitem[{Zhang et~al.(2022)Zhang, Li, Deng, Bing, and Lam}]{zhang2022survey}
Wenxuan Zhang, Xin Li, Yang Deng, Lidong Bing, and Wai Lam. 2022.
\newblock A survey on aspect-based sentiment analysis: tasks, methods, and
  challenges.
\newblock \emph{IEEE Transactions on Knowledge and Data Engineering}.

\bibitem[{Zhiltsova et~al.(2019)Zhiltsova, Caton, and
  Mulway}]{zhiltsova2019mitigation}
Alina Zhiltsova, Simon Caton, and Catherine Mulway. 2019.
\newblock Mitigation of unintended biases against non-native english texts in
  sentiment analysis.
\newblock In \emph{AICS}, pages 317--328.

\end{thebibliography}
\bibliographystyle{EMNLP}


\appendix

\section{Appendix}
\label{sec:appendix}

\begin{table*}[b]
\centering
\small
\begin{tabular}{|l|l|l|}
\hline
Term &
  Definition Framework &
  References \\ \hline
sentiment &
  affective state or feeling associated with a particular object or event & \cite{hoffmann2018too}
   \\ \hline
opinion &
  \begin{tabular}[c]{@{}l@{}}subjective statement, view, attitude, emotion, or appraisal about an entity\\ or an aspect of an entity from an opinion holder\end{tabular} & \cite{liu2012sentiment}
   \\ \hline
emotion/feelings &
  \begin{tabular}[c]{@{}l@{}}By “descriptive definition,” we mean a definition of the word emotion as\\  it is used in everyday life. By “prescriptive definition,” we mean a \\ definition of the scientific concept that is used to pick out the set of \\ events that a scientific theory of emotion purports to explain.\end{tabular} & \cite{izard2010many}
   \\ \hline
subjectivity &
  \begin{tabular}[c]{@{}l@{}} subjectivity analysis deals with the \\ detection of “private states” — a term that encloses sentiment, \\ opinions, emotions, evaluations, beliefs and speculations.\end{tabular} & \cite{wiebe1994tracking}
   \\ \hline
\end{tabular}
\caption{Examples of a few definitions of different themes concerning sentiment from different fields to demonstrate the difference in framework between these terms that are synonymously used in the field of SA in NLP.}
\end{table*}

\subsection{Application of SA}

\begin{figure*}[h]
  \centering
  \includegraphics[scale = 0.35]{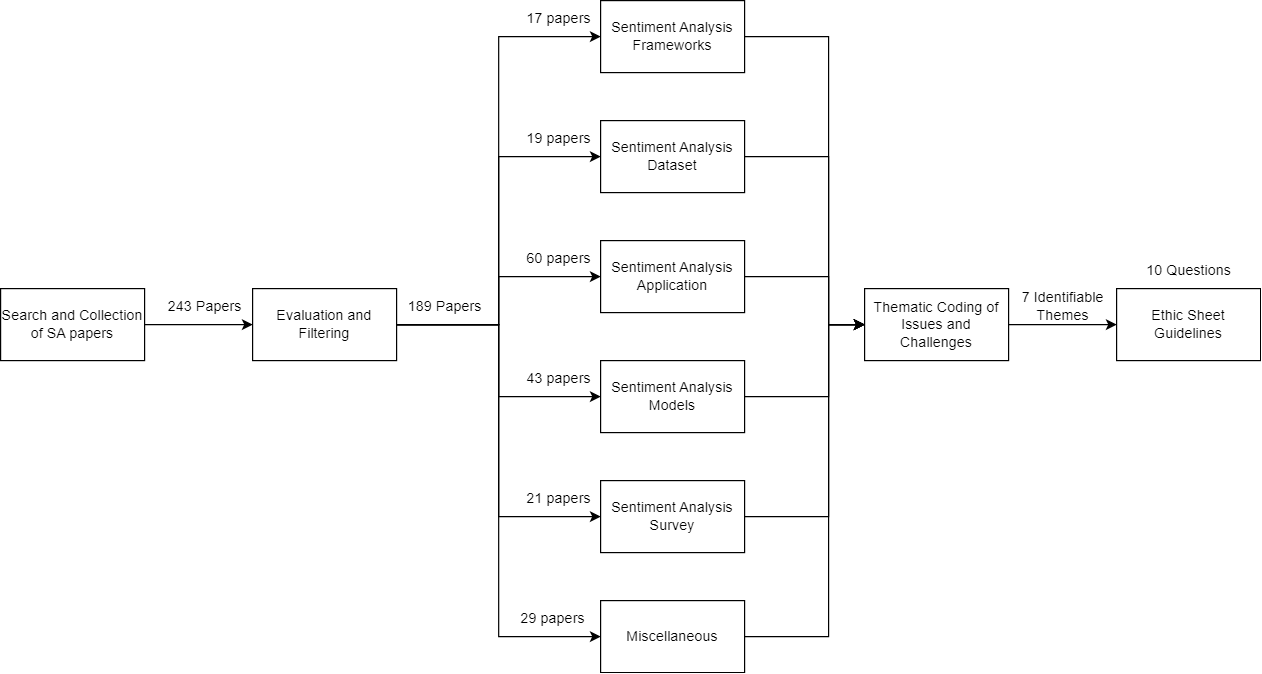}
  \caption{Roadmap of the collection and analysis process of all the peer-reviewed sentiment analysis papers to design the Ethics Sheet.}
  \label{fig:roadmap}
\end{figure*}

In this section, we illustrate the examples and categories of works that were looked into for understanding the various applications of SA. We categorize the purpose of SA into 5 major categories. The definitions and categories of all the applications are mentioned in Table \ref{table:application-groups}.

\textbf{Health and Medicine:} \citet{ji2013monitoring}, \citet{wu2015twitter}, \citet{rodrigues2016sentihealth}, \citet{bui2016temporal}, \citet{korkontzelos2016analysis}, \citet{asghar2016sentihealth}, \citet{du2017leveraging}, \citet{yang2016mining}, \citet{hassan2017sentiment}, \citet{ali2017sentiment}, \citet{gopalakrishnan2017patient}, \citet{birjali2017machine}, \citet{sabra2018prediction}, \citet{salas2017sentiment}, \citet{izzo201786},
\citet{crannell2016pattern}, \citet{rajput2020natural}, \citet{ramirez2019use}, \citet{wang2022global}, \citet{fang2023study}

\textbf{Government and Policy Making:}\citet{kwon2006multidimensional}, \citet{conrad2007opinion}, \citet{zavattaro2015sentiment}, \citet{yuliyanti2017sentiment}, 
\citet{syaifudin2017twitter},\citet{joyce2017sentiment}, \citet{shayaa2017social}, \citet{fatyanosa2017classification}, \citet{mansour2018social}, \citet{ikoro2018analyzing}, \citet{falck2020measuring}, \citet{georgiadou2020big}, \citet{ash2022measuring}

\textbf{Business Analytics:} \citet{fan2009blogger}, \citet{wang2013financial}, \citet{isah2014social}, \citet{akter2016sentiment}, \citet{li2016intelligent}, \citet{saragih2017sentiment}, \citet{jabbar2019real}, \citet{bose2020sentiment}, \citet{bonny2022sentiment}, \citet{li2023restaurant},  

\textbf{Social Media Analytics:} \citet{cao2013web, ortigosa2014sentiment, shahare2017sentiment, mahtab2018sentiment, abd2018sentiment, el2018novel, jing2019theoretical, rani2022efficient, venkit2021asourceful}

\textbf{Finance:} \citet{loughran2011liability, price2012earnings, schumaker2012evaluating, wang2013financial, garcia2013sentiment, rognone2020news, kraaijeveld2020predictive}

\subsection{Sentiment Analysis Models}
\citet{hu2004mining, wilson2005opinionfinder, sebastiani2006sentiwordnet, nielsen2011new, taboada2011lexicon, mohammad2013crowdsourcing, hannak2012tweetin, mohammad2012emotional, de2012pattern, wang2012system, gonccalves2013panas}
\citet{mohammad2013nrc, socher2013recursive, clement2013umigon, warriner2013norms, cambria2022guest, thelwall2014heart, hutto2014vader, gatti2015sentiwords, wang2016recursive, saeidi2016sentihood, baziotis2017datastories, moreno2018lingmotif, ma2018targeted}
\citet{deng2019sentiment, xu2019bert, sun2019utilizing, amplayo2019rethinking, rietzler2019adapt, lyu2020improving, wu2021context, cambria2022guest, ma2017cascading, devlin2018bert, liu2012sentiment, raffel2020exploring, yang2019xlnet, ionescu2019vector}
\citet{baccianella2010sentiwordnet, pappas2013distinguishing}

\subsection{Sentiment Analysis Datasets}
\citet{socher2013recursive, maas2011learning, wiebe2005annotating, li2018towards, barbieri2020tweeteval, rosenthal2019semeval, pang2004sentimental, pang2005seeing, nakov-etal-2013-semeval, barnes-etal-2022-semeval, alam2023gmnlp, blitzer2007biographies}
\citet{go2009twitter, misc_opinrank_review_dataset_205, majumder2019generating, he2016ups, alam2016joint, kiritchenko-mohammad-2018-examining}

\subsection{Sentiment Analysis Surveys}
\citet{medhat2014sentiment, alessia2015approaches, ribeiro2016sentibench, laskari2016aspect, zhang2018deep, zhang2022survey, sanchez2019social, drus2019sentiment, ramirez2019use, Kothari2020ANA, birjali2021comprehensive, mohammad2022ethics, guo2021overview}
\citet{zad2021survey, wankhade2022survey, soni2022survey, chan2023state}

\subsection{Bias in Sentiment Analysis}
\citet{huang_reducing_2020, diaz2018addressing, venkit2021identification, bhaskaran-bhallamudi-2019-good, kiritchenko-mohammad-2018-examining, zhiltsova2019mitigation, hube_debiasing_2020, han2018improving, sweeney2020reducing, prabhakaran-etal-2019-perturbation}
\citet{rozado2020wide, hutchinson-etal-2020-social, davidson2019racial, shen2018darling, narayanan2023unmasking, asyrofi2021biasfinder, ungless2023potential, lin2021mitigating, mei2023bias, venkit2023automated}

\subsection{Breakdown of the Metrics used in Sentiment Analysis}
\textbf{Sentiment Categorization}: Negative, Objective, Positive \cite{wilson2005opinionfinder}| Negative, Positive \cite{cambria2014senticnet}| Negative, Neutral, Positive \cite{wang2016recursive} | Very Negative, Negative, Neutral, Positive, Very Positive \cite{socher2013recursive} | Positive, Somewhat Positive, Neutral, Somewhat Negative, Negative \cite{devlin2018bert} | Valence, Arousal, Dominance \cite{warriner2013norms} | Negative, Neutral, Unsure, Positive \cite{de2012pattern}| Self-assurance, Attentiveness, Fatigue, Guilt, Fear, Sadness, Hostility, Joviality, Serenity, Suprise, Shyness \cite{gonccalves2013panas} | Joy, Sadness, Anger, Fear, Disgust, Surprise \cite{mohammad2012emotional}

\textbf{Sentiment Regression Scales}: [-5,+5] \cite{nielsen2011new}| [0,2,4] \cite{mohammad2013nrc}| [-1,+1] \cite{gonccalves2013panas} | [-4,+4] \cite{hutto2014vader}

\subsection{Breakdown of Ethics Sheet}

In this section, we aim to analyze the underlying intention behind each question posed in the Ethics Sheet.

Questions \textbf{(Q1-Q3)} are designed to focus on recommendation [R2]. The disclosure of all necessary information pertaining to the framework and analysis methodology is crucial. This disclosure contributes to the interpretability of sociotechnical systems employing SA, enhancing the understanding of their functioning.

Questions \textbf{(Q4-Q6)} are tailored to address recommendations [R1] and [R4]. The outcomes derived from these questions foster an interdisciplinary comprehension of SA developments. Explicitly stating user profiles and associated data empowers users with a democratic choice in selecting suitable applications as required.

Questions \textbf{(Q7-Q10)} emphasize the significance of comprehending weaknesses and biases inherent in a model. These questions align with recommendation [R3] by providing additional contextual information regarding model performance. The inclusion of information concerning implicit and explicit biases sheds light on the potential harm that a poorly administered model may exacerbate.

\end{document}